\documentclass{article}

\PassOptionsToPackage{numbers,compress}{natbib}
\usepackage[preprint]{neurips_2025}

\usepackage[utf8]{inputenc}
\usepackage[T1]{fontenc}
\usepackage{hyperref}
\usepackage{url}
\usepackage{booktabs}
\usepackage{amsfonts}
\usepackage{amsmath}
\usepackage{nicefrac}
\usepackage{microtype}
\usepackage{xcolor}
\usepackage{graphicx}
\usepackage{subcaption}
\usepackage{multirow}
\usepackage{cleveref}
\usepackage{pifont}
\usepackage{algorithm}
\usepackage{algpseudocode}
\usepackage{makecell}
\usepackage{array}
\usepackage{float}
\usepackage{listings}
\title{GUI-Perturbed: Domain Randomization Reveals Systematic Brittleness in GUI Grounding Models}

\author{
  Yangyue Wang\textsuperscript{1,2}
  \And Harshvardhan Sikka\textsuperscript{1,2}
  \And Yash Mathur\textsuperscript{* 2}
  \And Tony Zhou\textsuperscript{* 2}
  \AND Jinu Nyachhyon\textsuperscript{* 2}
  \And Pranav Guruprasad\textsuperscript{1,2}
}

\newcommand{\authorfootnote}{%
  \vspace{-2.0em}
  \begin{center}
    {\small \textsuperscript{*}Equal contributions. \;\;
    \textsuperscript{1}Fig \;\;
    \textsuperscript{2}Manifold Research}
  \end{center}
  \vspace{1.5em}
}

\begin{document}

\maketitle
\authorfootnote

\begin{abstract}

GUI grounding models report over 85\% accuracy on standard benchmarks, yet drop 27--56 percentage points when instructions require spatial reasoning rather than direct element naming. Current benchmarks miss this because they evaluate each screenshot once with a single fixed instruction. We introduce GUI-Perturbed, a controlled perturbation framework that independently varies visual scenes and instructions to measure grounding robustness. Evaluating three 7B models from the same architecture lineage, we find that relational instructions cause systematic accuracy collapse across all models, a 70\% browser zoom produces statistically significant degradation, and rank-8 LoRA fine-tuning with augmented data degrades performance rather than improving it. By perturbing along independent axes, GUI-Perturbed isolates which specific capability axes are affected---spatial reasoning, visual robustness, reasoning calibration---providing diagnostic signal that aggregate benchmarks cannot. We release the dataset, augmentation pipeline, and a fine-tuned model.
\end{abstract}


\section{Introduction}
\label{sec:intro}

GUI grounding, the task of locating a target element given a screenshot and a natural language instruction, is a fundamental capability for computer use agents (CUAs). Because every downstream action depends on first identifying the correct element, grounding errors compound throughout the entire interaction. State-of-the-art vision-language models now report over 85\% accuracy on ScreenSpot-v2~\citep{cheng2024seeclick} and up to 80.5\% on the more challenging ScreenSpot-Pro~\citep{li2025screenspotpro}, and these numbers increasingly inform deployment decisions for web automation and enterprise workflows.

These scores do not reflect deployment reliability. A model scoring 85\% on ScreenSpot-v2 can confuse a Google Sheets formula bar with a browser search bar, as both are white rectangles near the top of the screen. The same model's performance drops to 35\% when the instruction requires spatial reasoning (``click the button above X'') rather than direct naming (``click the Submit button''). A 70\% browser zoom is sufficient to further degrade it. We refer to this as the \emph{white rectangle problem}: models ground to visual primitives (shape, position, color) rather than functional semantics, and the resulting gap between benchmark performance and real-world robustness is systematic, not anecdotal.

Why do standard benchmarks miss this? Because they evaluate each screenshot exactly once with one fixed instruction~\citep{cheng2024seeclick, li2025screenspotpro, xie2024osworld, deng2023mind2web}. When the visual scene never changes, a model that has memorized where elements tend to appear looks indistinguishable from one that understands spatial layout. When instructions always name the target directly, a model with no spatial reasoning scores the same as one that can resolve ``the field above X.'' Benchmarks that introduce runtime anomalies~\citep{yang2025guirobust} or vary starting states~\citep{zhao2026worldgui} test complementary failure surfaces, but neither isolates which specific visual or instructional property caused a given failure.

The real world does not hold still. Browser zoom levels, font size preferences, website redesigns, and dark mode are variations that every user encounters. To measure robustness against them, we need an evaluation that perturbs visual and instructional conditions along controlled independent axes so that each failure can be traced to a specific change. We borrow this idea from domain randomization in robotics~\citep{tobin2017domain}, where textures, lighting, and colors are randomized in simulation to force policies to learn invariances. Training on fixed screenshots is the GUI equivalent of training on a single simulator skin.

To quantify these failures, we construct \textbf{GUI-Perturbed}, a controlled perturbation framework that applies domain randomization to GUI grounding evaluation. Using Mind2Web MHTML archives~\citep{deng2023mind2web} as our simulator and Playwright as the rendering engine, we perturb both the visual scene (style changes, zoom, text scaling) and the instruction (direct vs.\ spatial-relational) along independent axes. Evaluating three 7B models from the same Qwen2.5VL-7B~\citep{bai2025qwen25vl} lineage (Qwen2.5VL-7B, UI-TARS-1.5-7B~\citep{qin2025uitars}, and GTA1-7B~\citep{yang2025gta1}), we report the following findings:
\begin{enumerate}
  \item \textbf{Spatial reasoning is systematically deficient.} Relational instructions (``click the button above X'') cause 27--56\,pp accuracy collapse across all models. UI-TARS-1.5 drops 56.0\,pp despite scoring 89.7\% on ScreenSpot-v2.
  \item \textbf{Visual heuristics are static.} A 70\% browser zoom degrades all three models by 3--8~pp, with the largest effects on relational queries. Models encode absolute positions at a fixed scale rather than relational structure between elements.
  \item \textbf{The standard training recipe does not help.} Under rank-8 LoRA~\citep{hu2021lora}, no augmentation strategy improves performance on average, and scaling from 6.5k to 25k samples amplifies degradation on both GUI-Perturbed and ScreenSpot-v2. GUI-Perturbed exposes which spatial and visual axes are degrading, providing diagnostic granularity that standard benchmarks cannot.
\end{enumerate}

Additionally, both our baseline comparison and training experiments suggest that SFT with cross-entropy loss may degrade spatial reasoning. UI-TARS-1.5, trained on ${\sim}$50B GUI-focused tokens through SFT/DPO, achieves worse relational accuracy (35.0\%) than the base Qwen2.5VL (45.0\%) despite improving on direct grounding, and our rank-8 LoRA training experiments show a consistent pattern (\cref{sec:training}). GTA1, further trained with GRPO, recovers to 65.8\% relational accuracy but is the only model harmed by chain-of-thought reasoning across all conditions. Whether this reasoning sensitivity stems from the RL objective or from the coordinate-only output format used during training remains an open question (\cref{sec:results:reasoning}).

To support reproducibility and further research, we release:
\begin{itemize}
  \item \textbf{GUI-Perturbed}, a controlled perturbation benchmark (390 samples $\times$ 4 visual variants $\times$ 2 instruction types) for evaluating GUI grounding robustness.\footnote{\url{https://huggingface.co/datasets/figai/GUI-Perturbed}}
  \item \textbf{GUI-DR}, an open-source data augmentation pipeline for generating perturbation variants from MHTML archives, applicable to other datasets.\footnote{\url{https://github.com/ManifoldRG/GUI-DR}}
  \item \textbf{UI-TARS-1.5-7B-GUI-Perturbed}, a fine-tuned model, and result viewers for qualitative failure analysis.\footnote{\url{https://huggingface.co/figai/UI-TARS-1.5-7B-GUI-Perturbed}}
\end{itemize}

\section{Related Work}
\label{sec:related}

\subsection{GUI Grounding Benchmarks}
\label{sec:related:benchmarks}

\begin{table}[t]
\caption{Comparison of GUI grounding benchmarks. Scene variability: \emph{Fixed} = no visual variation; \emph{Live} = uncontrolled real website changes; \emph{Perturbed}/\emph{Systematic} = controlled variation deliberately introduced. GUI-Perturbed\dag\ is web-only; cross-platform extension is future work.}
\label{tab:benchmark-comparison}
\centering
\resizebox{\linewidth}{!}{%
\begin{tabular}{lllllll}
\toprule
\textbf{Data Paradigm} & \textbf{Dataset} & \textbf{Annotation Source} & \textbf{Platform} & \textbf{Size (Base Tasks)} & \textbf{Scene Variability} & \textbf{Variations per Task} \\
\midrule
\multirow{7}{*}{Fixed-scene}
 & OSWorld          & Human        & Desktop, Web, Mobile & 369    & Fixed & ---          \\
 & ScreenSpot-v2    & Human        & Desktop, Web, Mobile & 1,272  & Fixed & ---          \\
 & ScreenSpot-Pro   & Human        & Desktop              & 1,585  & Fixed & ---          \\
 & Mind2Web-2~\citep{gou2025mind2web2}       & Human        & Web                  & 130    & Fixed & ---          \\
 & VisualWebArena~\citep{koh2024visualwebarena}   & Human        & Web                  & 910    & Fixed & ---          \\
 & MiniWoB++~\citep{shi2017miniwob}        & Programmatic & Web (simulated)      & 100+   & Fixed & ---          \\
 & OSWorld-G~\citep{xie2025jedi}        & Human + LLM  & Desktop              & 564    & Fixed & ---          \\
\midrule
Live-scene & Online-Mind2Web  & Human        & Web                  & 300    & Live  & ---          \\
\midrule
\multirow{3}{*}{Perturbation-based}
 & GUI-Robust       & Human + MLLM & Desktop, Web         & 5,318  & Perturbed & Single-axis (Anomalies) \\
 & WorldGUI         & Human        & Desktop              & 315    & Perturbed & Single-axis (Initial state) \\
 & \textbf{GUI-Perturbed\dag} & \textbf{Programmatic} & \textbf{Web} & \textbf{3,120 (390$\times$8)} & \textbf{Systematic} & \textbf{Multi-axis (visual \& inst.)} \\
\bottomrule
\end{tabular}%
}
\end{table}

Fixed-scene benchmarks such as OSWorld~\citep{xie2024osworld}, ScreenSpot-v2~\citep{cheng2024seeclick}, ScreenSpot-Pro~\citep{li2025screenspotpro}, and Mind2Web~\citep{deng2023mind2web} all share the same assumption: one screenshot, one instruction, no variation (\cref{tab:benchmark-comparison}). Several recent efforts address aspects of this limitation. GUI-Robust~\citep{yang2025guirobust} tests robustness under runtime anomalies such as pop-ups and error dialogs, which is complementary to GUI-Perturbed's focus on pre-action visual and instruction variation. WorldGUI~\citep{zhao2026worldgui} varies the starting states of desktop environments but cannot isolate which specific visual property caused a failure. Live benchmarks such as Online-Mind2Web~\citep{xue2025onlinemind2web} exhibit natural scene variation, but because that variation is uncontrolled, failures cannot be attributed to specific visual changes. GUI-Perturbed is the only benchmark that perturbs both the visual scene and the instruction along controlled, independent axes, enabling attribution of each failure to a specific perturbation type.

\subsection{Domain Randomization}
\label{sec:related:dr}

Domain randomization is a standard sim-to-real transfer technique in robotics~\citep{tobin2017domain}, in which textures, lighting, and colors are randomized during training so that the policy learns invariances rather than memorizing a single environment. The parallel to GUI agents is direct: training on fixed screenshots is analogous to training on a single simulator skin. Since GUI environments lack programmatic visual control, we use MHTML archives as our simulator, manipulating the DOM and re-rendering via Playwright to produce controlled visual variation.

\subsection{GUI Agent Training Data}
\label{sec:related:training}

Building robust GUI agents also requires appropriate training data. Real trajectory collection remains expensive~\citep{wang2025opencua, qin2025uitars}, and synthetic generation is fragile~\citep{xie2025jedi}, still requiring substantial real data in the training mix. Beyond collection cost, existing training recipes organize data by surface features (platform, action type, element type) rather than by the cognitive capabilities they exercise. Our training experiments in \cref{sec:training} investigate whether augmentation data generated through domain randomization can address this gap.

\section{The GUI-Perturbed Framework}
\label{sec:framework}

\subsection{Design Principles}
\label{sec:framework:principles}

We isolate step-level grounding: given a single screenshot and a natural language instruction, the model must predict the target element. This formulation removes multi-step dependencies so that each failure is directly attributable to grounding. We build on Mind2Web MHTML archives~\citep{deng2023mind2web}, which provide DOM-level access for semantically meaningful perturbations rather than pixel-level transforms.

\subsection{Two Axes of Perturbation}
\label{sec:framework:axes}

A grounding model takes two inputs: a visual scene (the screenshot) and an instruction (the natural language description of the target element). We perturb both. \emph{Visual scene perturbations} alter the rendered page while preserving the target element, changing the visual context in which the model must locate the target. \emph{Instruction perturbations} vary the referring expression from direct (naming the target by its text or type) to relational (identifying the target by its spatial relationship to a landmark).

\subsection{Perturbation Variants}
\label{sec:framework:variants}

\begin{table}[t]
  \caption{GUI-Perturbed dataset statistics. Each variant contains the same 390 grounding steps with matched ground-truth annotations. Instruction types are evaluated independently per variant.}
  \label{tab:dataset-stats}
  \centering
  \small
  \begin{tabular}{llc}
    \toprule
    Variant & Description & Eval Samples \\
    \midrule
    Original & Mind2Web MHTML rendered via Playwright & 390 \\
    Style & Randomized button orders, element styles via CSS/JS & 390 \\
    Precision & Page scaled to $0.7\times$ (70\% zoom) & 390 \\
    Text Shrink & Reduced text font size & 390 \\
    \midrule
    \multicolumn{2}{l}{Instruction types per variant} & Direct + Relational \\
    \multicolumn{2}{l}{Reasoning modes per configuration} & With CoT + Without CoT \\
    \multicolumn{2}{l}{Total evaluation configurations per model} & $4 \times 2 \times 2 = 16$ \\
    \bottomrule
  \end{tabular}
\end{table}

\Cref{tab:dataset-stats} summarizes the dataset. Each variant contains 390 grounding steps derived from Mind2Web interaction traces. The \textbf{Original} variant renders MHTML files directly via Playwright without modification. The \textbf{Style} variant randomizes button orders and element styles through injected CSS/JS. The \textbf{Precision} variant scales the page to $0.7\times$ (70\% zoom), simulating a common browser zoom setting. The \textbf{Text Shrink} variant reduces text font sizes while preserving layout structure.

\subsection{Relational Instructions}
\label{sec:framework:relational}

Instead of naming the target directly, relational instructions identify it by spatial relationship to a landmark: ``above,'' ``below,'' ``to the left of,'' ``to the right of.'' This reflects natural human referring behavior. Combined with visual perturbations, relational instructions test whether models maintain structured spatial representations or rely on memorized co-occurrences. The generation procedure is formalized in \cref{alg:gui-gen} (\textsc{FindRelationalAnchor} and \textsc{GenerateInstructions}).

\subsection{Data Generation Pipeline (GUI-DR)}
\label{sec:framework:pipeline}

The pipeline is open-source and proceeds as follows: given a Mind2Web step record and a perturbation variant, we render the MHTML archive via Playwright, apply the specified visual perturbation, re-locate the target bounding box, generate both direct and relational instructions, and capture the final screenshot. \Cref{alg:gui-gen} formalizes the entire procedure.

\begin{algorithm}[t]
\caption{GUI-Perturbed Dataset Generation (One Step)}
\label{alg:gui-gen}
\begin{algorithmic}[1]
\Statex \textbf{Input:}\quad Mind2Web step record (action $a$, target element description $e$, original bbox $b$), MHTML path, variant $v \in \{\mathsf{original},\ \mathsf{style},\ \mathsf{precision},\ \mathsf{text\_shrink}\}$
\Statex \textbf{Output:}\quad Trajectory entry $(\mathsf{img},\ b'_{\text{updated bbox}},\ \mathsf{instr}_{\rm direct},\ \mathsf{instr}_{\rm relational})$
\Statex

\Function{ApplyVariant}{$\text{browser},\ v$}
    \If{$v = \mathsf{style}$}
        \State $\theta \gets \text{Sample}(\{\mathsf{neobrutalism},\ \mathsf{glassmorphism},\ \ldots\})$
        \State $\text{InjectStylesheet}(\text{browser},\ \theta)$;\quad $\text{ShuffleDOM}(\text{browser})$
    \ElsIf{$v = \mathsf{precision}$}
        \State $\text{ScaleViewport}(\text{browser},\ 0.7)$
    \ElsIf{$v = \mathsf{text\_shrink}$}
        \State $\text{SetFontSizes}(\text{browser},\ f \leftarrow \max(0.8\,f,\ 11))$;\quad $\text{RelaxOverflow}(\text{browser})$
    \EndIf
\EndFunction

\Statex

\Function{FindRelationalAnchor}{$e_{\text{target}}$}
    \State $e_{\text{anchor}} \gets \text{NearestInteractable}(e_{\text{target}})$ \Comment{closest element to target}
    \State $\mathsf{dir} \gets \text{SpatialDirection}(e_{\text{anchor}},\ e_{\text{target}})$ \Comment{$\in \{\mathsf{above},\ \mathsf{below},\ \mathsf{left},\ \mathsf{right}\}$}
    \State \Return $(e_{\text{anchor}},\ \mathsf{dir})$
\EndFunction

\Statex

\Function{GenerateInstructions}{$a,\ e_{\text{target}},\ e_{\text{anchor}},\ \mathsf{dir}$}
    \State $\mathsf{instr}_{\rm direct} \gets \text{Template}(a,\ e_{\text{target}})$ \Comment{e.g., ``Click on `Submit' button''}
    \State $\mathsf{instr}_{\rm relational} \gets \text{Template}(a,\ e_{\text{anchor}},\ \mathsf{dir})$ \Comment{e.g., ``Click on the button above `Email'\,''}
    \State \Return $(\mathsf{instr}_{\rm direct},\ \mathsf{instr}_{\rm relational})$
\EndFunction

\Statex

\Function{GUIPerturbedGeneration}{$a,\ e_{\text{target}},\ b,\ \text{MHTML},\ v$}
    \State $\text{browser} \gets \text{RenderMHTML}(\text{MHTML})$
    \State $\text{ApplyVariant}(\text{browser},\ v)$
    \State $b' \gets \text{LocateBbox}(\text{browser},\ b,\ e_{\text{target}})$
    \State $(e_{\text{anchor}},\ \mathsf{dir}) \gets \text{FindRelationalAnchor}(e_{\text{target}})$
    \State $(\mathsf{instr}_{\rm direct},\ \mathsf{instr}_{\rm relational}) \gets \text{GenerateInstructions}(a,\ e_{\text{target}},\ e_{\text{anchor}},\ \mathsf{dir})$
    \State $\mathsf{img} \gets \text{CaptureScreenshot}(\text{browser})$
    \State \Return $(\mathsf{img},\ b',\ \mathsf{instr}_{\rm direct},\ \mathsf{instr}_{\rm relational})$
\EndFunction

\end{algorithmic}
\end{algorithm}

\paragraph{Training data filtering.} We filter via rejection sampling using Holo2-30B-A3B as a teacher model, which scores 66.1 on ScreenSpot-Pro.

\paragraph{Evaluation data filtering.} We conduct a manual review of each sample step, rejecting it unless all 4 variants pass the following criteria:
\begin{enumerate}
  \item The target element bounding box is correct.
  \item The bounding box is centered on the target element.
  \item The ground truth element text and surrounding context are realistic.
  \item The UI is not extremely unrealistic (slightly occluded elements are acceptable).
  \item The instruction is unambiguous for the target element (text matches, element type matches, no duplicate targets).
\end{enumerate}

The pipeline is released as a reusable augmentation tool~\citep{gui_perturbed_technical_report_2026}.

\begin{table}[t]
\centering
\caption{Training data splits used in \cref{sec:training}. These splits are generated via the GUI-DR pipeline.}
\label{tab:data-splits}
\small
\begin{tabular}{lll}
\toprule
\textbf{Data Split} & \textbf{Variant Composition} & \textbf{Sample Size} \\
\midrule
6.5k style & style & 6,500 \\
6.5k text shrink precision & text shrink + precision & 6,500 \\
6.5k all & style + text shrink + precision & 6,500 \\
25k all & 1 original + 5 style + 1 text shrink + 1 precision & 24,935 \\
\bottomrule
\end{tabular}
\end{table}

\paragraph{Intended usage.} GUI-Perturbed can serve as a complementary robustness benchmark alongside standard evaluations such as ScreenSpot-v2. The GUI-DR pipeline can be applied to other MHTML-based datasets to generate perturbation variants for training or evaluation. We release evaluation scripts and a result viewer for qualitative failure analysis. The current dataset is built on Mind2Web and covers web-only scenarios; we welcome community contributions to expand coverage to other domains (desktop, mobile) and data sources. We note that while the evaluation data underwent manual filtering (\cref{sec:framework:pipeline}), edge cases may remain, and we encourage community feedback to improve data quality over time.

\section{Experimental Setup}
\label{sec:setup}

\subsection{The Triple Alignment Problem}
\label{sec:setup:alignment}

GUI grounding requires simultaneous alignment along three axes: visual alignment (recognizing element appearance), functional alignment (understanding element affordance), and geometric alignment (reasoning about spatial relations between elements). Standard benchmarks evaluate these capabilities in an entangled manner. GUI-Perturbed is designed to isolate the visual and geometric axes independently.

\subsection{POMDP Formulation}
\label{sec:setup:pomdp}

We formulate the CUA setting as a POMDP $\mathcal{M} = \langle S, A, O, T(s_{t+1} \mid s_t, a_t), R \rangle$. In the general multi-step setting, the agent receives an instruction $I$ and at each step $t$ observes $O_t$ (a screenshot sampled from the hidden state via $O_t \sim Z(O_t \mid s_t)$), optionally produces a chain-of-thought trace $t_t$, and outputs an action $a_t$. The action is conditioned on the full trajectory history: all prior observations $O_{1:t}$, reasoning traces $t_{1:t-1}$, and actions $a_{1:t-1}$:
\begin{equation}
  (t_t, a_t) = \text{VLM}_\theta(I, O_{1:t}, t_{1:t-1}, a_{1:t-1})
\end{equation}
The observation decomposes along three alignment axes: $O_t \supseteq (O_t^{\text{vis}}, O_t^{\text{geo}}, O_t^{\text{func}})$, corresponding to visual appearance, geometric layout, and functional affordance respectively.

Our evaluation isolates the single-step grounding case: given a single instruction $I$ and a single observation $O$, predict the correct element. This is equivalent to evaluating the first step of the trajectory ($t{=}1$) with no prior history. This removes multi-step dependencies so that every failure is unambiguously a grounding failure. In multi-step tasks, failures are ambiguous across instruction understanding, element grounding, and action selection. Isolating grounding eliminates this confound.

\subsection{Models}
\label{sec:setup:models}

\subsubsection{Open Models (Controlled Comparison)}

We evaluate three 7B models sharing the Qwen2.5VL-7B~\citep{bai2025qwen25vl} base, differing only in post-training (\cref{tab:model-comparison,tab:model-benchmarks}). All three share the same architecture and base weights, so any performance difference is attributable solely to post-training. This controlled setup enables us to isolate the effect of each successive stage of GUI-specialized training on robustness under perturbation. Each model is evaluated in both reasoning and no-reasoning configurations. Qwen2.5-VL and UI-TARS-1.5 use their native prompt formats; GTA1, which has no native reasoning template, uses a reasoning prompt adapted from UI-TARS-1.5. Full prompt templates are provided in \cref{sec:appendix:prompts}.

\begin{table}[t]
\centering
\renewcommand{\arraystretch}{2.0}
\caption{Model architecture and training. All three models share the Qwen2.5VL-7B base. Differences arise solely from post-training recipe and data.}
\label{tab:model-comparison}
\resizebox{\textwidth}{!}{%
\begin{tabular}{>{\bfseries}p{2.2cm} p{3.5cm} p{6cm} p{5cm}}
\toprule
\textbf{Model} & \textbf{Architecture} & \textbf{Training Stages} & \textbf{Data} \\
\midrule
Qwen2.5VL-7B
& Qwen2.5VL-7B
& \makecell[tl]{
  \textbf{3 stages:}\\
  \textbf{1.} Visual Pre-Training (1.5T tokens)\\
  \textbf{2.} Multimodal Pre-Training (2T tokens)\\
  \textbf{3.} Long-Context Pre-Training\\
  \quad (0.6T tokens) + SFT + DPO
}
& \makecell[tl]{
  \textbf{4.1T tokens total:}\\
  $\bullet$ Interleaved image-text VQA\\
  $\bullet$ Image Captions \& OCR\\
  $\bullet$ Visual knowledge\\
  $\bullet$ Video Grounding\\
  $\bullet$ Document parsing\\
  $\bullet$ Agent interaction data
} \\
\midrule
UI-TARS1.5-7B
& Qwen2.5VL-7B
& \makecell[tl]{
  \textbf{3 stages from UI-TARS recipe}\\
  (UI-TARS1.5 recipe not open source):\\
  \textbf{1.} Continual Pre-training\\
  \quad (GUI interaction knowledge)\\
  \textbf{2.} Annealing Phase (UI-TARS-SFT)\\
  \textbf{3.} DPO Phase (UI-TARS-DPO)
}
& \makecell[tl]{
  \textbf{$\sim$50B tokens:}\\
  $\bullet$ 18.4M grounding elements\\
  \quad (web / mobile / desktop)\\
  $\bullet$ 6M GUI tutorials\\
  $\bullet$ 151.4k action traces\\
  $\bullet$ Reflective online traces
} \\
\midrule
GTA1-7B
& \makecell[tl]{
  UI-TARS1.5-7B\\
  (grounding model)\\
  + o3 planner\\
  (test-time scaling)
}
& \makecell[tl]{
  \textbf{RL Optimization:}\\
  $\bullet$ GRPO (Group Relative\\
  \quad Policy Optimization)\\
  $\bullet$ Click reward mechanism
}
& \makecell[tl]{
  $\bullet$ Aria-UI~\citep{ariaui}\\
  $\bullet$ OmniAct~\citep{kapoor2024omniact}\\
  $\bullet$ Widget Caption~\citep{li2020widget}\\
  $\bullet$ UI-Vision~\citep{nayak2025uivisiondesktopcentricguibenchmark}\\
  $\bullet$ OS-Atlas~\citep{wu2024osatlas}\\
  (lightly cleaned)
} \\
\bottomrule
\end{tabular}%
}
\end{table}

\begin{table}[t]
\centering
\caption{Published benchmark scores for the three evaluated models. Scores are reported from \citet{yang2025gta1}.}
\label{tab:model-benchmarks}
\small
\begin{tabular}{lccccc}
\toprule
\textbf{Model} & \textbf{ScreenSpot-v2} & \textbf{ScreenSpot-Pro} & \textbf{OSWorld} & \textbf{OSWorld-G} \\
\midrule
Qwen2.5VL-7B & 88.8 & 27.6 & -- & 27.7 \\
UI-TARS1.5-7B & 89.7 & 42.0 & 27.4$\pm$2.2\% & 64.2 \\
GTA1-7B & \textbf{92.4} & \textbf{50.1} & \textbf{45.2} (with o3) & \textbf{67.7} \\
\bottomrule
\end{tabular}
\end{table}

\subsection{Evaluation Configuration}
\label{sec:setup:eval}

4 visual variants $\times$ 2 instruction types $\times$ 2 reasoning modes $=$ 16 configurations per open model.

Given a predicted point $\hat{p}_i = (\hat{x}_i, \hat{y}_i)$ and ground-truth bounding box $b_i = (x_i, y_i, w_i, h_i)$ with center $p_i = (x_i + w_i/2,\; y_i + h_i/2)$, we evaluate perturbation robustness along three complementary dimensions, each computed over $n{=}390$ matched sample pairs per perturbation test (same task and step evaluated on both the original and perturbed screenshots)

\begin{itemize}
  \item \textbf{Hit rate}: the proportion of predictions that fall inside the ground-truth bounding box. We report 95\% bootstrap confidence intervals (10,000 resamples) for all hit rates; exact binomial (Clopper--Pearson) intervals agreed within 0.2\,pp throughout and are omitted for brevity.
  \item \textbf{Flip rate}: the fraction of matched pairs whose binary outcome (hit/miss) changed between the original and the perturbed condition. This measures \emph{prediction consistency}: a high flip rate indicates the model's output is sensitive to the perturbation, regardless of whether accuracy improves or degrades on average.
  \item \textbf{Net~$\Delta$}: the difference in hit rate between the original and perturbed conditions (original $-$ perturbed), with 95\% bootstrap CI. A positive $\Delta$ indicates degradation. We test significance with McNemar's test, which compares the number of samples that degraded ($b$: correct $\to$ incorrect) against those that improved ($c$: incorrect $\to$ correct) under perturbation, ignoring samples whose outcome did not change. We report $p$-values with continuity correction; when the number of discordant pairs ($b + c$) is below 25, we use the exact binomial test instead. Each model is tested in 4 configurations (2 instruction types $\times$ 2 reasoning modes); the ``Sig.'' column in \cref{tab:robustness-baseline} reports how many of these 4 tests reached $p < 0.05$.
\end{itemize}

Flip rate and net~$\Delta$ decompose the perturbation effect: a perturbation can cause many individual predictions to change (high flip rate) without shifting overall accuracy (low $\Delta$), if roughly equal numbers of samples degrade and improve. Conversely, a perturbation with a high $\Delta$ necessarily has a high flip rate with an asymmetric split between degraded and improved samples. We additionally report bounding box center MSE, normalized MSE, and normalized distance in \cref{sec:appendix:metrics}, though these did not reveal trends beyond what hit rate already captures.

\section{Results}
\label{sec:results}
We evaluate the three baseline models under perturbation~\citep{measuring_gui_models_robustness_technical_report_2026}.
\begin{table}[t]
\centering
\caption{Perturbation robustness of baseline models ($n = 390$ matched sample pairs per test).}
\label{tab:robustness-baseline}
\small
\begin{tabular}{llc cc cc cc}
\toprule
 & & & \multicolumn{2}{c}{\textbf{Flip Rate}} & \multicolumn{2}{c}{\textbf{Net $\Delta$ (\%)}} & & \\
\cmidrule(lr){4-5} \cmidrule(lr){6-7}
\textbf{Model} & \textbf{Pert.} & \textbf{Base Acc.} & \textbf{Dir.} & \textbf{Rel.} & \textbf{Dir.} & \textbf{Rel.} & $\boldsymbol{b}$\,/\,$\boldsymbol{c}$ & \textbf{Sig.} \\
\midrule
GTA-1 & Precision & 79.3 & 10.3\% & 21.5\% & +3.6** & +7.9*** & 169/79 & 3/4 \\
 & Style &  & 9.7\% & 21.5\% & +1.3 & -0.3 & 126/118 & 0/4 \\
 & Text Shrink &  & 4.2\% & 16.7\% & +0.4 & +1.8 & 90/73 & 0/4 \\
\midrule
Qwen2.5-VL & Precision & 66.0 & 13.1\% & 16.4\% & +3.3 & +4.9* & 147/83 & 2/4 \\
 & Style &  & 8.7\% & 19.1\% & +2.8** & +0.1 & 120/97 & 1/4 \\
 & Text Shrink &  & 7.8\% & 14.4\% & +0.1 & +2.8 & 98/75 & 0/4 \\
\midrule
UI-TARS-1.5 & Precision & 63.0 & 13.1\% & 18.7\% & +6.2*** & +5.4** & 169/79 & 4/4 \\
 & Style &  & 11.2\% & 19.2\% & +2.4 & +1.0 & 132/105 & 0/4 \\
 & Text Shrink &  & 6.9\% & 14.1\% & -0.8 & +0.0 & 79/85 & 0/4 \\
\bottomrule
\end{tabular}
\vspace{4pt}
\begin{minipage}{\linewidth}
\scriptsize
Base Acc.\ = hit rate (\%) on unperturbed screenshots, averaged across reasoning modes and query types.
Flip Rate = fraction of matched pairs whose outcome changed. Dir.\ = direct instructions; Rel.\ = relational.
Net\,$\Delta$ = hit-rate drop (pp); positive = degradation.
$b$/$c$ = samples degraded/improved, aggregated across configurations.
Sig.\ = significant McNemar tests ($p < 0.05$) out of 4.
*/~**/~*** = $p < 0.05$/$0.01$/$0.001$.
\end{minipage}
\end{table}

\begin{figure}[t]
  \centering
  \includegraphics[width=\linewidth]{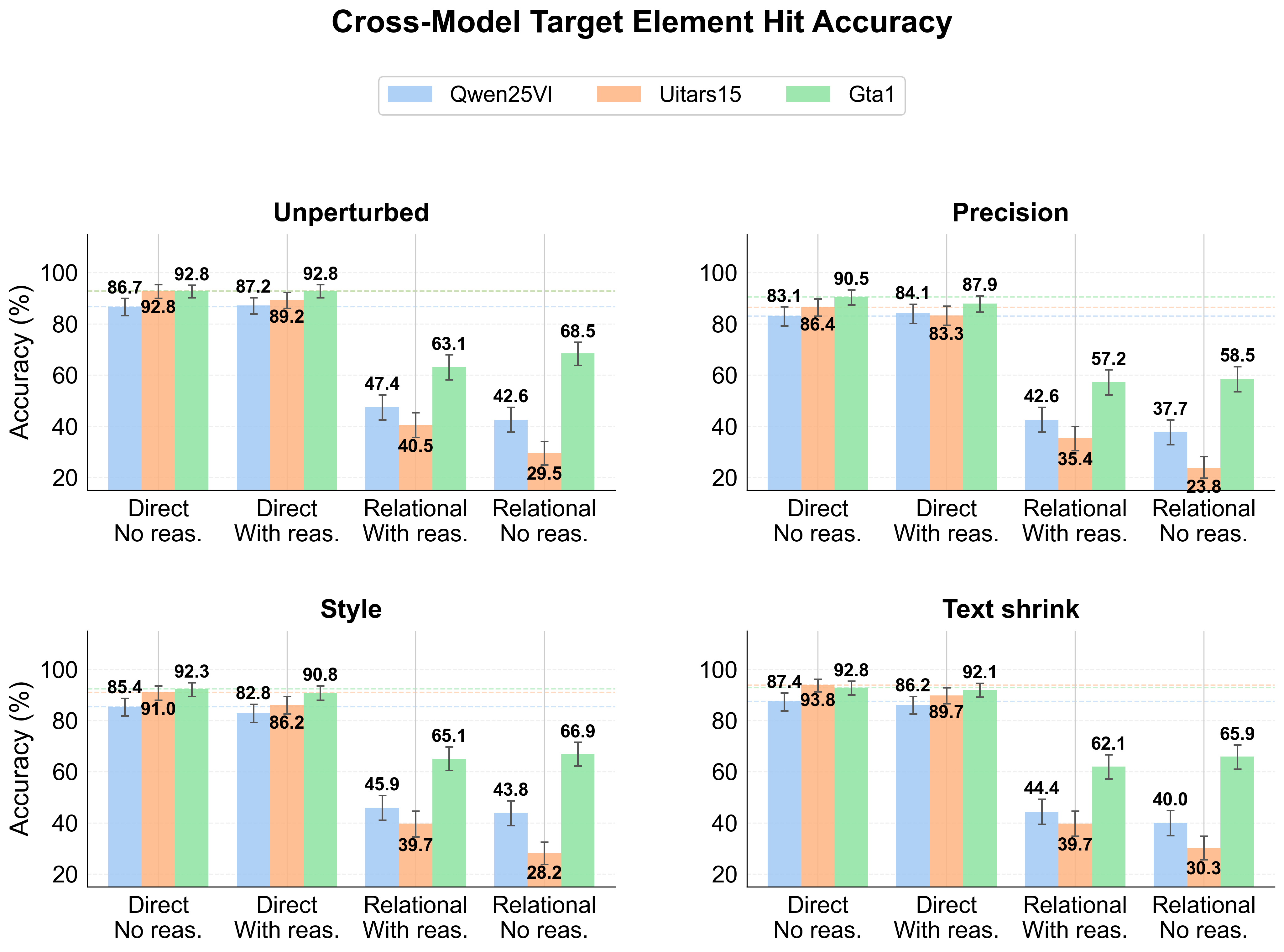}
  \caption{Hit rates with 95\% bootstrap confidence intervals across models and configurations.}
  \label{fig:baseline-hitrate-ci}
\end{figure}

\begin{figure}[t]
  \centering
  \includegraphics[width=\linewidth]{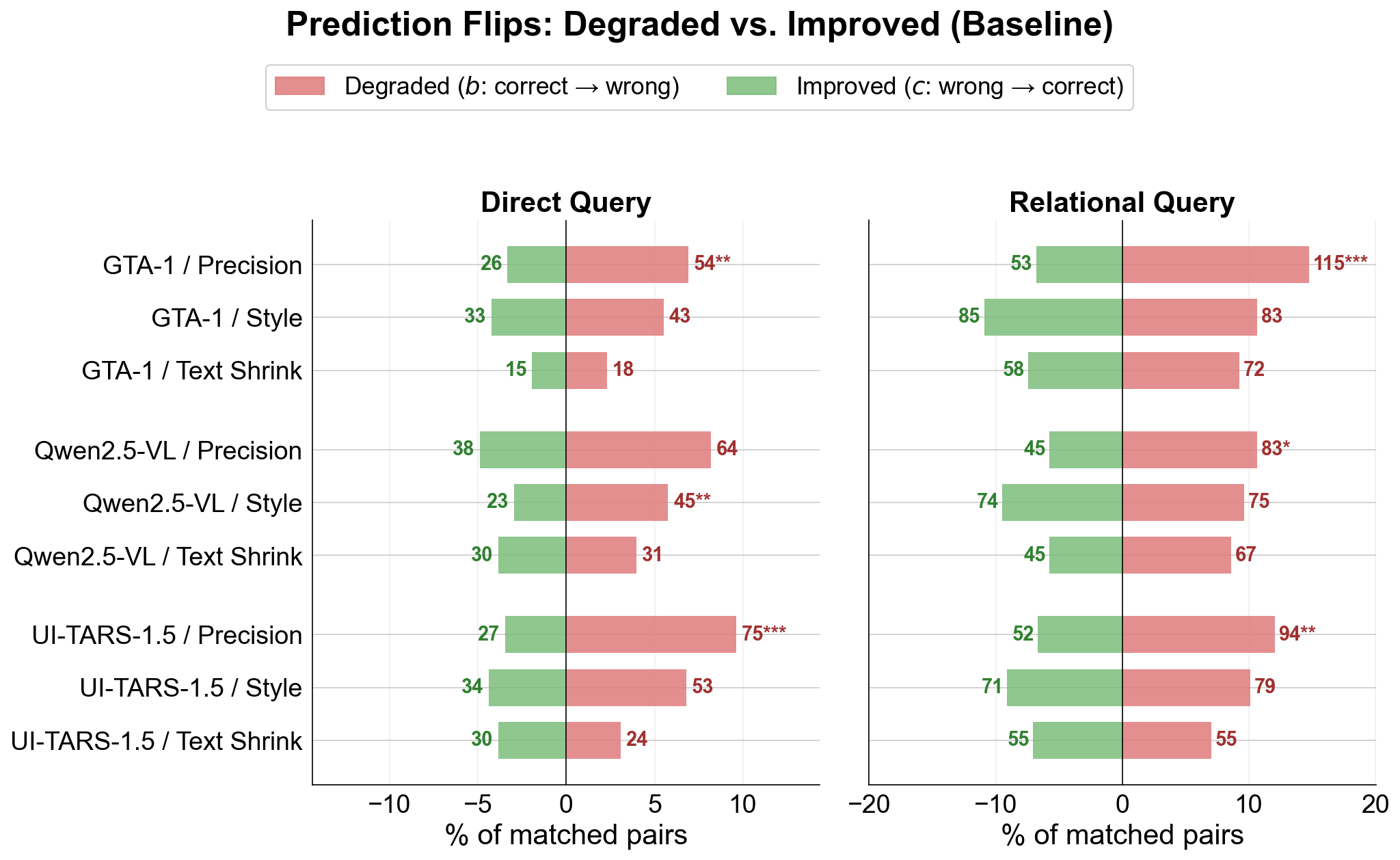}
  \caption{Flip rate decomposition for baseline models under each perturbation type.}
  \label{fig:baseline-flip}
\end{figure}

\subsection{Visual Perturbations Degrade High-Scoring Models}
\label{sec:results:visual}

Precision perturbation (70\% zoom) produced statistically significant accuracy drops in 9 of 12 paired comparisons (McNemar's $p < 0.05$), compared to 1/12 for style and 0/12 for text-shrink (\cref{tab:robustness-baseline}). Aggregating $b$/$c$ counts across all 12 configurations (3 models $\times$ 2 instruction types $\times$ 2 reasoning modes), the effect was consistently unidirectional: 485 predictions flipped from correct to incorrect ($b$) versus 241 that improved ($c$), a $b$/$c$ ratio of $\approx$ 2.0:1. By contrast, style perturbation produced a near-symmetric split (378 degraded vs.\ 320 improved, 1.2:1).

The following per-model results compare hit rates on the original (unperturbed) variant against the precision variant, averaged across reasoning modes. The effect was most pronounced on relational queries:

\begin{itemize}
  \item \textbf{GTA-1}: direct 92.8\% (95\% CI [90.3, 95.3]) $\to$ 89.2\% (CI [86.0, 92.2]), drop 3.6\,pp ($p = 0.006$); relational 65.8\% (CI [61.0, 70.4]) $\to$ 57.8\% (CI [52.9, 62.7]), drop 7.9\,pp ($p < 0.001$).
  \item \textbf{UI-TARS-1.5}: direct 91.0\% (CI [88.1, 93.8]) $\to$ 84.9\% (CI [81.3, 88.3]), drop 6.2\,pp ($p < 0.001$); relational 35.0\% (CI [30.4, 39.7]) $\to$ 29.6\% (CI [25.0, 34.2]), drop 5.4\,pp ($p = 0.008$). UI-TARS-1.5 was the only model where precision perturbation was significant in all 4 configurations (4/4).
  \item \textbf{Qwen2.5-VL}: direct 86.9\% (CI [83.6, 90.0]) $\to$ 83.6\% (CI [79.9, 87.2]), drop 3.3\,pp ($p = 0.071$, n.s.); relational 45.0\% (CI [40.1, 49.9]) $\to$ 40.1\% (CI [35.3, 45.0]), drop 4.9\,pp ($p = 0.023$).
\end{itemize}

Models encode absolute spatial positions at a fixed scale rather than relational structure between elements. A zoom change shifts pixel positions enough to break these memorized associations. In the triple alignment framework (\cref{sec:setup:alignment}), this constitutes a \emph{visual alignment} failure: the model's learned representations are too tightly coupled to the specific pixel-level statistics of the training distribution.

Importantly, the lack of significance for style and text-shrink perturbations does not mean these perturbations had no effect on individual predictions (\cref{fig:baseline-flip,fig:baseline-hitrate-ci}). Aggregating across all 12 configurations (3 models $\times$ 2 instruction types $\times$ 2 reasoning modes, 390 matched pairs each), style perturbation flipped 698 of 4,680 predictions (14.9\% flip rate), comparable to precision's 726 flips (15.5\%). However, because style flips were roughly bidirectional (378 degraded vs.\ 320 improved), the net accuracy change was not statistically distinguishable from zero. This reveals a distinction between \emph{robustness} (net $\Delta$) and \emph{consistency} (flip rate): all three perturbation types cause substantial prediction instability, but only precision perturbation does so in a systematically harmful direction.

Some perturbations unexpectedly improved accuracy on individual configurations (e.g., style on relational+CoT for GTA-1: 63.1\% $\to$ 65.1\%, +2.1\,pp), though none reached significance (all $p > 0.4$). Similar non-significant improvements appeared for text-shrink on UI-TARS-1.5 direct queries (92.8\% $\to$ 93.8\%, $p = 0.45$).

\subsection{Spatial Reasoning as the Primary Failure Mode}
\label{sec:results:spatial}

Relational instructions caused 27--56\,pp accuracy drops compared to direct instructions. All comparisons were significant at $p < 0.001$ (two-proportion $z$-test):

\begin{itemize}
  \item \textbf{GTA-1}: 92.8\% (95\% CI [90.3, 95.3]) $\to$ 65.8\% (CI [61.0, 70.4]), drop 27.1\,pp ($z = 8.61$--$10.02$, $p < 0.001$).
  \item \textbf{Qwen2.5-VL}: 86.9\% (CI [83.6, 90.0]) $\to$ 45.0\% (CI [40.1, 49.9]), drop 41.9\,pp ($z = 11.83$--$12.88$, $p < 0.001$).
  \item \textbf{UI-TARS-1.5}: 91.0\% (CI [88.1, 93.8]) $\to$ 35.0\% (CI [30.4, 39.7]), drop 56.0\,pp ($z = 14.25$--$18.15$, $p < 0.001$).
\end{itemize}

\begin{figure}[t]
  \centering
  \includegraphics[width=\linewidth]{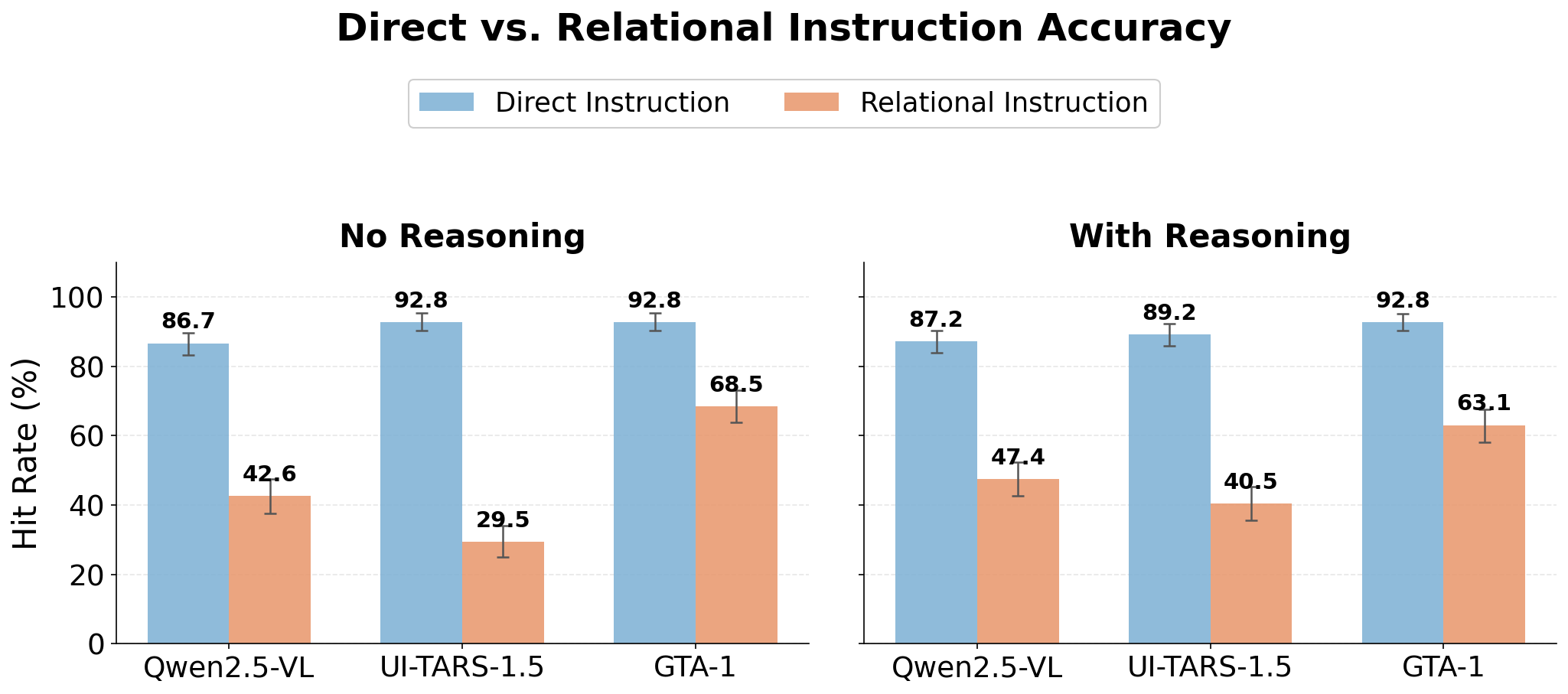}
  \caption{Direct vs.\ relational instruction accuracy across models. The 95\% bootstrap CIs do not overlap for any model.}
  \label{fig:direct-vs-relational}
\end{figure}

The 95\% bootstrap CIs for direct and relational hit rates do not overlap for any model (\cref{fig:direct-vs-relational}), confirming these are fundamental capability gaps rather than marginal differences. \Cref{tab:cross-benchmark} contextualizes these results against established benchmarks.

\begin{table}[t]
  \caption{Cross-benchmark comparison. GUI-Perturbed results are on the original (unperturbed) variant, averaged across reasoning modes. Relational instructions expose accuracy gaps not visible on standard benchmarks. Bold indicates best score per row.}
  \label{tab:cross-benchmark}
  \centering
  \small
  \begin{tabular}{lccc}
    \toprule
    Benchmark & Qwen2.5VL-7B & UI-TARS1.5-7B & \textbf{GTA1-7B} \\
    \midrule
    ScreenSpot-v2 & 88.8 & 89.7 & \textbf{92.4} \\
    ScreenSpot-Pro & 27.6 & 42.0 & \textbf{50.1} \\
    OSWorld & --- & 27.4$\pm$2.2 & \textbf{45.2} \\
    OSWorld-G & 27.7 & 64.2 & \textbf{67.7} \\
    \midrule
    GP-Unperturbed (Direct) & 86.9 & 91.0 & \textbf{92.8} \\
    GP-Unperturbed (Relational) & 45.0 (\textcolor{red}{$\downarrow$41.9}) & 35.0 (\textcolor{red}{$\downarrow$56.0}) & \textbf{65.8} (\textcolor{red}{$\downarrow$27.1}) \\
    \bottomrule
  \end{tabular}
\end{table}

The effect is consistent across reasoning modes. UI-TARS-1.5 shows the largest gap: 63.3~pp drop without CoT versus 48.7~pp with CoT. Chain-of-thought partially mitigates this deficit but does not resolve it.

We note that this is not a data quantity problem: these models were trained on millions of screenshots, yet still fail on relational instructions. Notably, the partial recovery observed with chain-of-thought suggests that the visual information needed to resolve spatial relations is present in the image; the models fail to extract it without explicit reasoning scaffolding. We discuss potential root causes in \cref{sec:discussion:spatial}.

\begin{figure}[t]
  \centering
  \includegraphics[width=\linewidth]{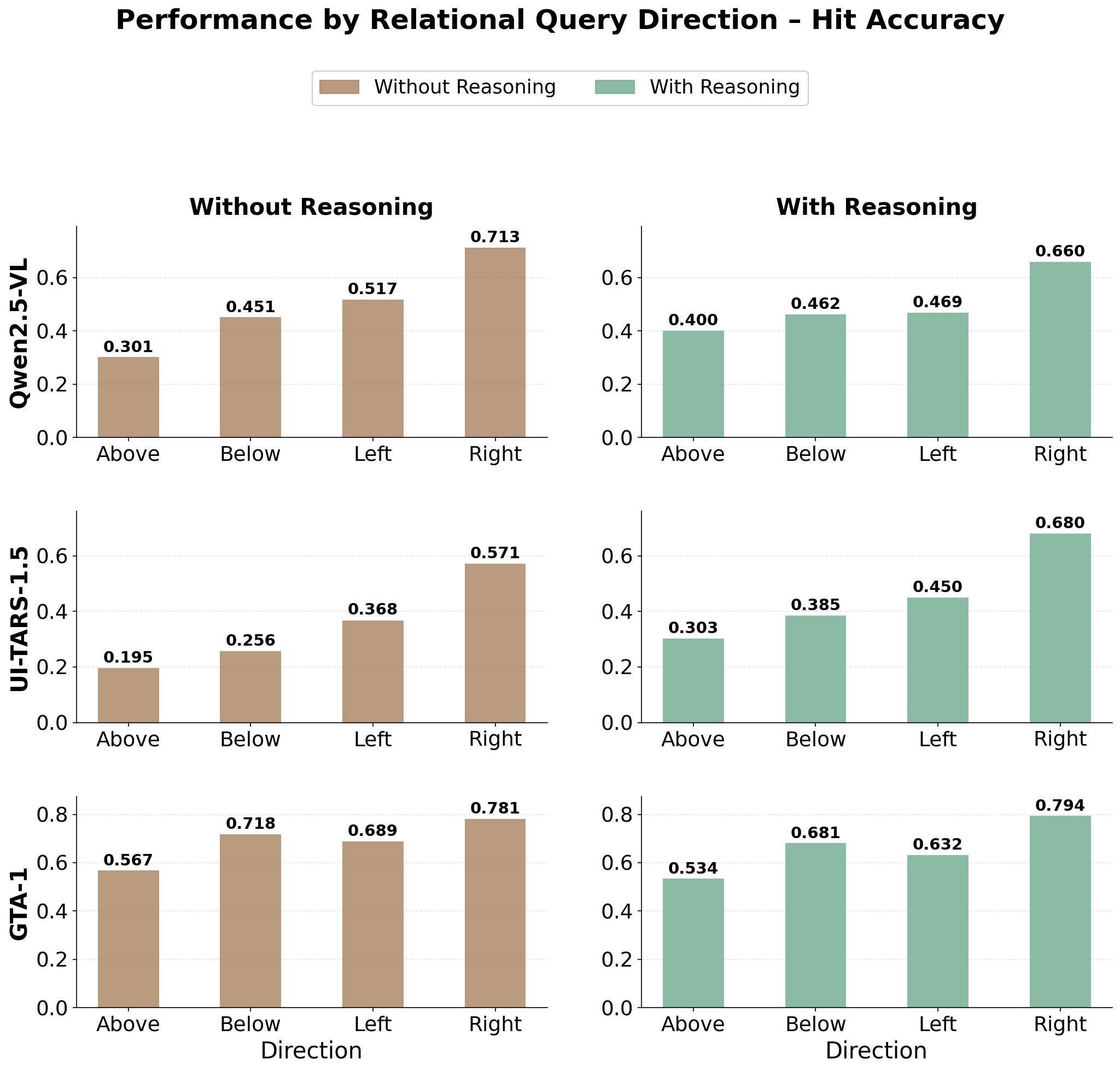}
  \caption{Accuracy by spatial direction (above, below, left, right) across models.}
  \label{fig:directional-bias}
\end{figure}

\paragraph{Directional bias.} We observe that models achieve higher accuracy on ``right'' instructions compared to ``left'' or ``above'' instructions (\cref{fig:directional-bias}), which may reflect biases in the training data distribution or the visual patchification order. Confirming the precise cause requires a larger controlled study.

\subsection{Effect of Chain-of-Thought Reasoning}
\label{sec:results:reasoning}

The effect of reasoning is not uniformly positive. On direct grounding tasks, enabling chain-of-thought (CoT) introduces unnecessary deliberation that can actively mislead the final prediction: the model overthinks a task that base visual grounding would handle correctly without intermediate reasoning. On relational tasks, CoT recovers performance by providing useful intermediate structure, allowing the model to reason about spatial relationships step by step rather than resolving them in a single forward pass.

GTA1 presents a particularly instructive case. Having been further trained with GRPO for direct coordinate prediction, it achieves the highest robustness on relational tasks among all models (65.8\% vs.\ 45.0\% and 35.0\%), yet it is harmed by CoT across \emph{all} conditions, including relational tasks where the other two models benefit from reasoning. Whether this sensitivity stems from the RL objective itself or from the coordinate-only output format used during training (which may overfit the model away from CoT capability) remains an open question.

The implication is that uniformly enabling reasoning is not the right strategy. Models need exposure to diverse reasoning styles during post-training, and the optimal reasoning style and length likely varies by task. GUI-Perturbed can track both the spatial robustness and reasoning sensitivity effects, making it a useful diagnostic tool for evaluating post-training recipes.

\subsection{Failure Mode Taxonomy}
\label{sec:results:taxonomy}

We conducted a qualitative analysis of representative failures across all models and configurations. \Cref{tab:failure-modes} summarizes the recurring failure modes, organized into four categories. The qualitative analysis reveals distinctions within failure categories that aggregate metrics cannot capture.

\paragraph{Spatial failures} are the most prevalent category, with three distinct mechanisms:
\begin{itemize}
  \item \emph{Click region errors}: the model identifies the correct element but clicks the wrong physical area, indicating imprecise coordinate prediction.
  \item \emph{Location hallucinations}: the model names the correct element in its reasoning but outputs fabricated coordinates, indicating a disconnect between reasoning and action.
  \item \emph{Spatial reasoning errors}: the model incorrectly interprets directional relations (left/right, above/below), indicating failure in relational understanding itself.
\end{itemize}
These three modes have different implications: click region errors may be addressable through coordinate refinement, while spatial reasoning errors require representational changes.

\paragraph{Semantic failures} reveal grounding shortcuts:
\begin{itemize}
  \item \emph{Text matching bias}: the model clicks a visible text match without verifying it is the correct UI element (e.g., clicking a ``First Name'' label rather than the input field below it), revealing over-reliance on lexical matching.
  \item \emph{Goal hallucination}: the model invents user intent absent from the instruction, suggesting that the language prior can override the visual grounding signal.
  \item \emph{Instruction misinterpretation}: the model selects a related but incorrect element, misunderstanding what the instruction refers to.
\end{itemize}

\paragraph{Visual failure.} In \emph{visual confusion}, the model relies on superficial visual cues (shape, color, position) and misidentifies the functional element. For example, a model may mistake a light-colored button for a search box because both share similar visual properties.

Qualitative examples for each mode are provided in \cref{sec:appendix:taxonomy}.

\begin{table}[t]
  \caption{Failure mode taxonomy derived from qualitative analysis across all models and configurations.}
  \label{tab:failure-modes}
  \centering
  \small
  \begin{tabular}{lll}
    \toprule
    Category & Failure Mode & Description \\
    \midrule
    \multirow{3}{*}{Spatial}
      & Click Region Error & Correct element identified, wrong physical area clicked \\
      & Location Hallucination & Correct element named, fabricated coordinates output \\
      & Spatial Reasoning Error & Incorrect interpretation of above/below/left/right \\
    \midrule
    \multirow{3}{*}{Semantic}
      & Goal Hallucination & Model invents intent not present in instruction \\
      & Instruction Misinterpretation & Related but incorrect element selected \\
      & Text Matching Bias & Clicks visible text match without proper grounding \\
    \midrule
    Visual & Visual Confusion & Reliance on shape/color/position heuristics \\
    \midrule
    Reasoning & Reasoning Drift & CoT misleads final action prediction \\
    \bottomrule
  \end{tabular}
\end{table}

\section{Training Experiments}
\label{sec:training}

We investigate whether post-training can address the failures identified in \cref{sec:results}. We fine-tune with rank-8 LoRA SFT, a resource-efficient low-rank adaptation configuration. The training does not close the identified gaps (\cref{sec:training:insight})~\citep{training_on_gui_perturbed_technical_report_2026}.

\subsection{Setup}
\label{sec:training:setup}

\paragraph{Base model.} We fine-tune UI-TARS-1.5-7B, directly connecting to the evaluation gaps identified in \cref{sec:results}.

\paragraph{Training method.} We use LoRA~\citep{hu2021lora} at rank 8 (0.042\% trainable parameters), a resource-efficient configuration. See \cref{sec:limitations} for discussion of scope.

\paragraph{Data.} We prepare two training datasets at matched scale. The \emph{GUI-Perturbed training split} is synthetic data generated via GUI-DR on the Mind2Web training set and filtered with Holo2-30B-A3B, producing 24,935 steps across 8 variants (\cref{tab:data-splits}). The \emph{Salesforce GUI Grounding mix} serves as a real-data baseline, consisting of 25k samples uniformly sampled from five open-source grounding datasets: Aria-UI~\citep{ariaui}, OmniAct~\citep{kapoor2024omniact}, Widget Caption~\citep{li2020widget}, UI-Vision~\citep{nayak2025uivisiondesktopcentricguibenchmark}, and OS-Atlas~\citep{wu2024osatlas}. This pairing enables a controlled comparison between synthetic targeted data and real diverse data at the same scale.

\paragraph{Experiments.} We design three experiments to isolate different factors in the training pipeline. \emph{Experiment~1 (perturbation type)} tests which augmentation variants are most effective by comparing style-only, text-shrink+precision, and all-combined training sets, each at 6.5k samples. \emph{Experiment~2 (data scale)} tests whether more augmentation data improves robustness by comparing 6.5k and 25k samples of the all-combined variant. \emph{Experiment~3 (data source)} compares the Salesforce real-data mix (25k) against GUI-Perturbed synthetic data (25k) to determine whether the bottleneck is data quality or the training recipe. All models are evaluated on both GUI-Perturbed and ScreenSpot-v2.

\subsection{Perturbation Type Effects}
\label{sec:training:types}

\begin{figure}[t]
  \centering
  \includegraphics[width=\linewidth]{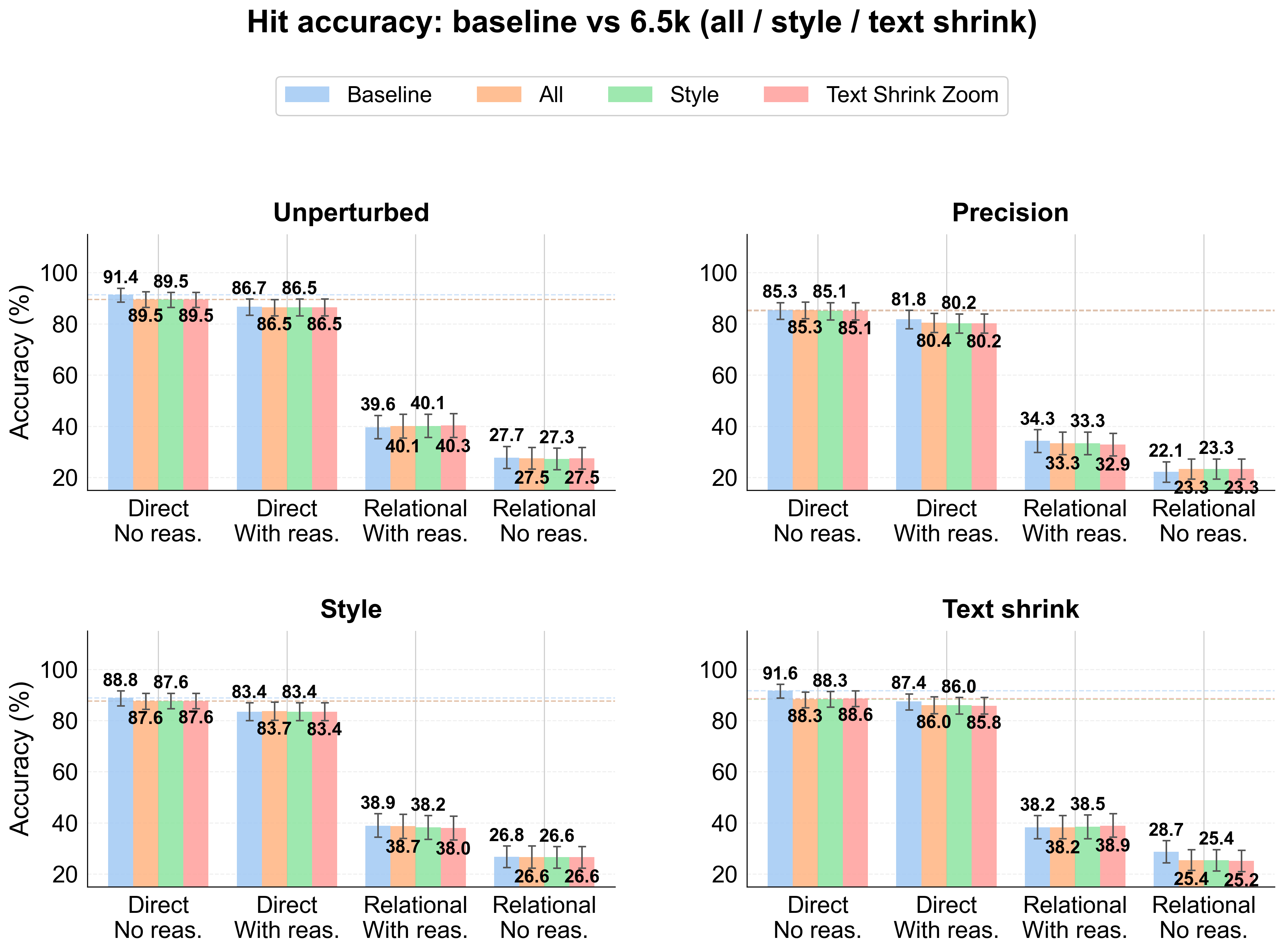}
  \caption{Effect of different augmentation types on model performance. All variants cause slight degradation; text shrink+precision is worst (${\sim}3.3$~pp on direct + no-reasoning).}
  \label{fig:ft-augmentation}
\end{figure}

\begin{figure}[t]
  \centering
  \includegraphics[width=\linewidth]{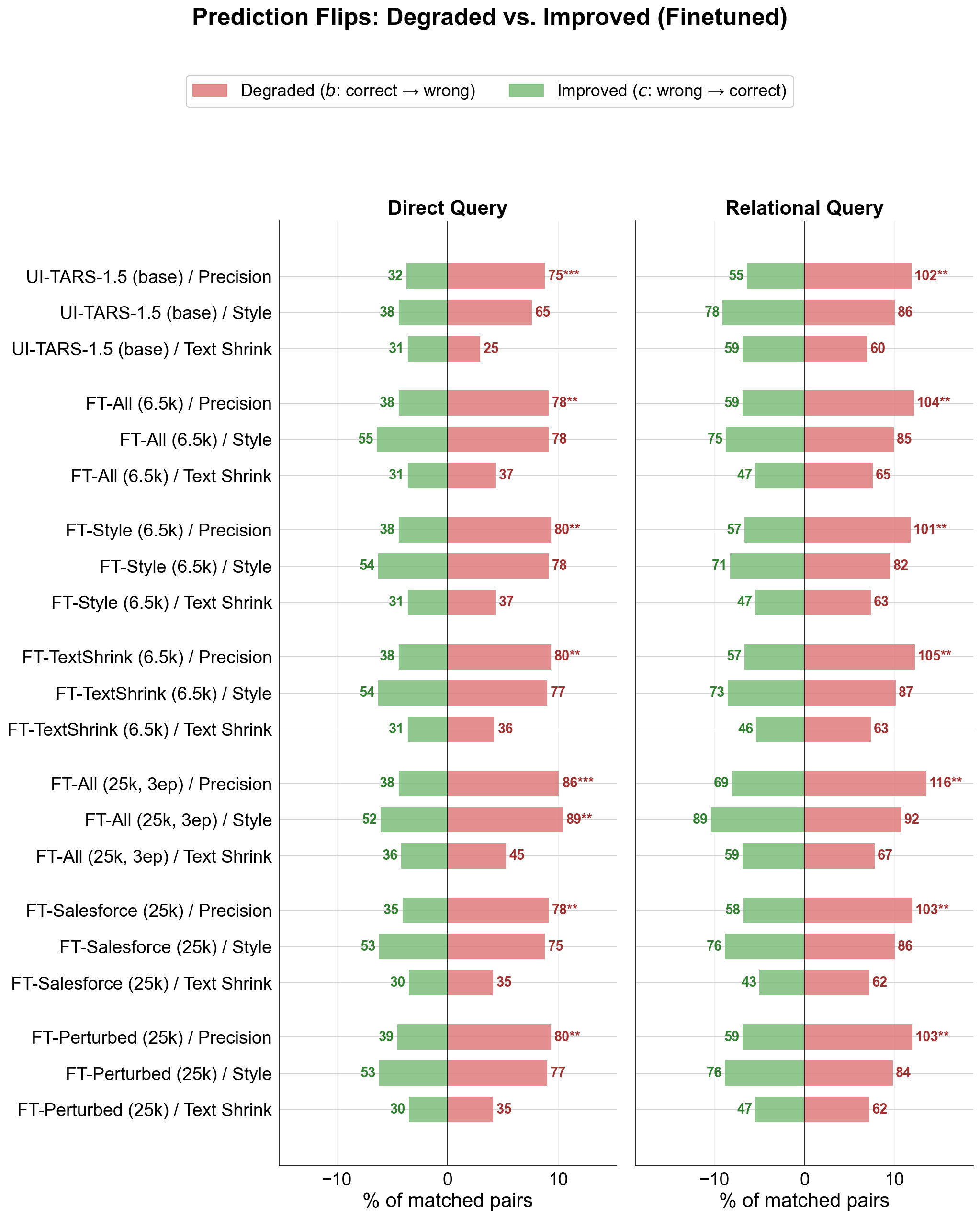}
  \caption{Flip rate decomposition for fine-tuned models under each augmentation type.}
  \label{fig:ft-flip}
\end{figure}

None of the augmentation variants improve performance on average (\cref{fig:ft-augmentation,fig:ft-flip}). Although individual configurations show flat or marginal changes, the overall trend is degradation. The text shrink+precision variant produces the largest drop (${\sim}3.3$~pp on direct queries without reasoning). This is notable because text shrink would be expected to be the gentlest perturbation; however, the associated changes in layout and font scale introduce a distribution shift that rank-8 LoRA cannot absorb, causing the model to fit perturbation artifacts rather than learn scale invariance. The complete robustness statistics for all finetuned variants are reported in \cref{tab:robustness-finetuned} (Appendix).

\subsection{Scaling Amplifies Degradation}
\label{sec:training:scaling}

\begin{figure}[t]
  \centering
  \includegraphics[width=\linewidth]{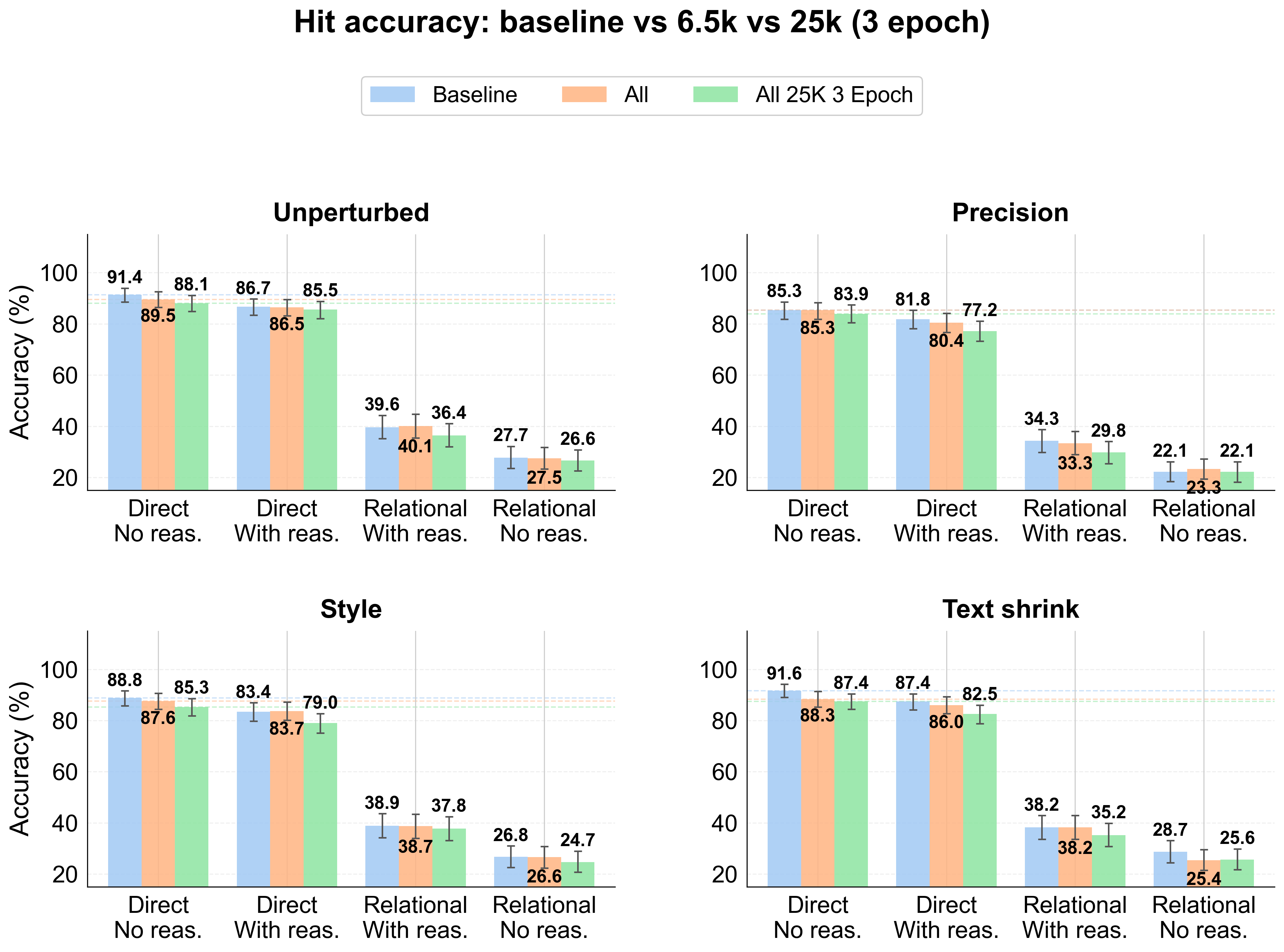}
  \caption{Effect of scaling training data from 6.5k to 25k samples. More data leads to worse performance on both GUI-Perturbed and ScreenSpot-v2.}
  \label{fig:ft-scaling}
\end{figure}

Scaling from 6.5k to 25k samples degrades performance on both GUI-Perturbed and ScreenSpot-v2 (\cref{fig:ft-scaling}). Two factors interact: catastrophic forgetting as distribution shift compounds with more data, and rank-8 LoRA memorizing perturbation artifacts rather than learning the intended invariances.

\subsection{Real vs.\ Synthetic Data}
\label{sec:training:realsynth}

\begin{figure}[t]
  \centering
  \includegraphics[width=\linewidth]{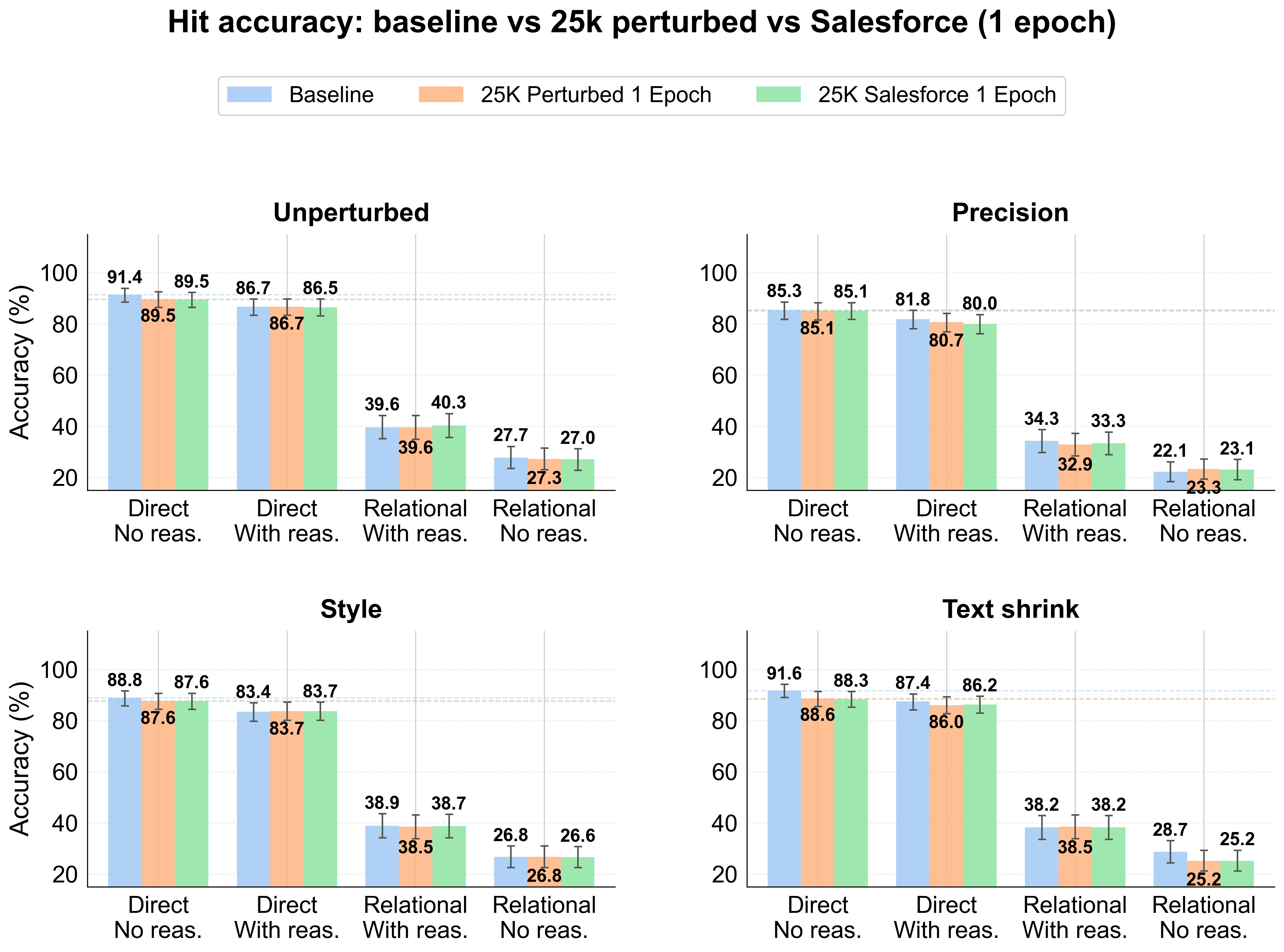}
  \caption{Comparison of real (Salesforce mix) vs.\ synthetic (GUI-Perturbed) training data. Neither improves performance.}
  \label{fig:ft-realsynth}
\end{figure}

Neither real data (Salesforce mix) nor synthetic data (GUI-Perturbed) improves performance (\cref{fig:ft-realsynth}); both degrade the model, though in different ways. Real data degrades uniformly across perturbation types, while synthetic data shows larger drops on the specific perturbation types it was trained on, suggesting the model overfits to perturbation artifacts rather than learning invariance. The bottleneck appears to be the training recipe rather than the data source, consistent with the baseline finding that SFT/DPO-trained UI-TARS-1.5 degrades on relational tasks despite improving on direct grounding.

\subsection{Standard Benchmarks Mask Training Failures}
\label{sec:training:insight}

\begin{figure}[htbp]
  \centering
  \includegraphics[width=0.85\linewidth, height=0.4\textheight, keepaspectratio]{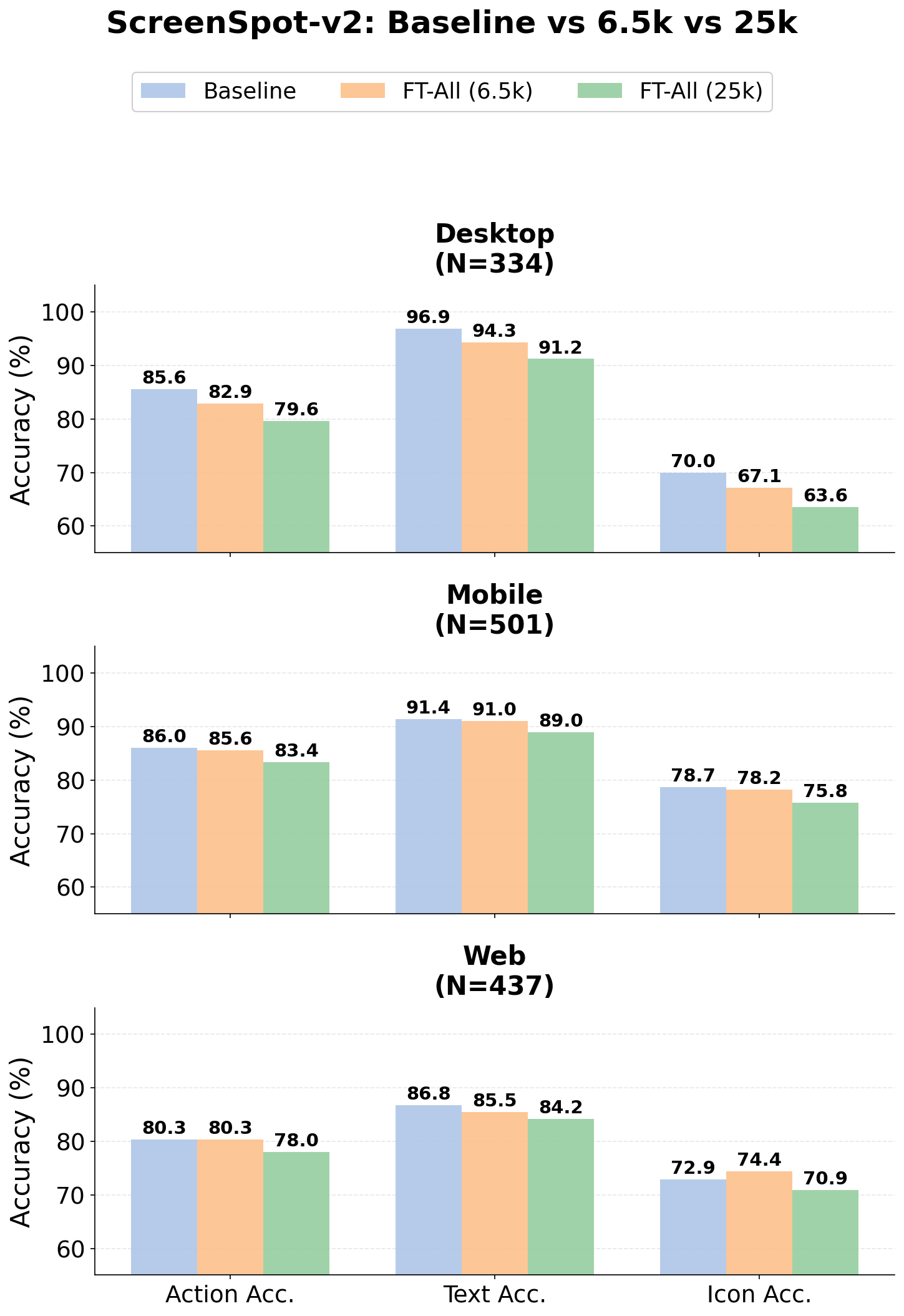}
  \caption{ScreenSpot-v2 accuracy by platform and element type: Baseline vs.\ FT-All (6.5k) vs.\ FT-All (25k). Scaling from 6.5k to 25k amplifies degradation across all categories.}
  \label{fig:benchmark-mask-1}
\end{figure}

As shown in \cref{fig:benchmark-mask-1}, the degradation patterns from Sections~\ref{sec:training:types}--\ref{sec:training:realsynth} are consistent across all training configurations. Standard benchmarks detect overall degradation from these interventions but cannot isolate which robustness axes are affected. For example, UI-TARS-1.5 drops approximately 6~pp on ScreenSpot-v2 desktop action accuracy when scaled from baseline to 25k training samples (\cref{fig:benchmark-mask-1}). GUI-Perturbed reveals that this degradation is concentrated on precision-perturbed direct queries (85.3\% $\to$ 77.2\%, drop of 8.1~pp) while precision-perturbed relational queries with reasoning degrade by 4.5~pp (34.3\% $\to$ 29.8\%), a per-axis breakdown invisible to aggregate evaluation.

\section{Discussion}
\label{sec:discussion}

\subsection{GUI Models Lack Spatial Relational Understanding}
\label{sec:discussion:spatial}

These models were trained on millions of screenshots yet still fail on relational instructions such as ``above X,'' indicating that additional data volume alone does not resolve this limitation. The root cause may lie in architecture (patch-level ViT encoders produce visual tokens without explicit spatial structure), in training (cross-entropy SFT provides no direct gradient signal for spatial reasoning), in positional encoding (current schemes may not capture inter-element spatial relations with sufficient fidelity), in the absence of targeted spatial reasoning data, or in some combination of these factors. Our experiments cannot isolate which, but they establish that the failure is consistent across models with different post-training recipes.

The problem may be particularly acute in GUI grounding because GUI elements are visually similar (buttons, text fields, and links share shapes and colors) and are often distinguished primarily by spatial context rather than appearance. The partial recovery we observe with chain-of-thought (\cref{sec:results:spatial}) suggests that the visual information needed for spatial reasoning is present in the image, but current models cannot extract it without explicit step-by-step reasoning. In our triple alignment framing (\cref{sec:setup:alignment}), this constitutes a geometric alignment failure.

\subsection{Visual Heuristics Are Static and Fragile}
\label{sec:discussion:heuristics}

Models memorize position-based associations that degrade under layout or style changes. In our zoom perturbation results, we observe models clicking on elements that occupy the spatial position where the target appeared at the original scale, indicating that the model memorized a position rather than learned a function. This is consistent with findings from \citet{yu2026visualattributes}, who observe that GUI agents are disproportionately affected by changes to visual properties that should be semantically irrelevant.

The deployment implications are significant. Websites routinely update their designs, run A/B tests with different layouts, and ship seasonal themes. A CUA built on static visual heuristics is one redesign away from failure. In our triple alignment framing, this is a visual alignment problem: the model's learned representations are coupled to the specific pixel-level statistics of the training distribution rather than to the functional properties of GUI elements.

\subsection{Reasoning Is Not Uniformly Beneficial}
\label{sec:discussion:reasoning}

The CoT results in \cref{sec:results:reasoning} have a broader implication for how CUAs are deployed and evaluated. Current deployment practice typically treats reasoning as a binary switch: enable or disable CoT globally. Our results indicate that this is insufficient. The interaction between reasoning mode, task type, and post-training recipe produces a three-way dependency that global reasoning policies cannot capture. Evaluation frameworks need to measure reasoning effects per task type, and post-training pipelines need to expose models to varied reasoning styles and lengths so they can learn when deliberation helps and when it hurts.

\subsection{Limitations of Low-Rank Adaptation for Spatial Grounding}
\label{sec:discussion:training}

Our training experiments test LoRA SFT with cross-entropy loss, the standard recipe for lightweight post-training of VLMs. Under this recipe, data distribution matters more than scale: small amounts of misaligned data cause disproportionate degradation because low-rank updates have limited capacity and allocate it to fitting whatever signal is strongest in the training distribution. The practical implication is that practitioners who collect or generate more GUI data without carefully controlling its distributional properties may find their models degrade rather than improve. Additionally, the loss does not directly supervise spatial reasoning. Cross-entropy optimizes next-token prediction over the action output but provides no gradient signal for the spatial representations that produce the correct action. Consequently, a model can learn to output plausible coordinates without improving its internal spatial representations. This observation is consistent with findings from DoRA~\citep{liu2024dora}, GLAD~\citep{peng2025glad}, and EvoCUA~\citep{xue2026evocua}, all of which report that LoRA fine-tuning of vision-language models can degrade capabilities in unexpected ways.

Our baseline evaluation provides additional evidence that SFT alone is insufficient for geometric understanding. UI-TARS-1.5, trained on ${\sim}$50B GUI-focused tokens through SFT/DPO, achieves \emph{worse} relational accuracy (35.0\%) than the base Qwen2.5VL (45.0\%), despite improving on direct grounding. GTA1, which adds GRPO with step-level click reward on top of UI-TARS-1.5, recovers to 65.8\%. This progression suggests that supervised fine-tuning on GUI data can improve direct element matching while degrading spatial reasoning, and that reinforcement learning with grounding-specific reward is more effective at teaching geometric understanding. Our rank-8 LoRA training experiments are consistent with this pattern: SFT with cross-entropy loss does not produce the representational changes that spatial reasoning requires.

A broader implication of our training experiments is methodological. Without perturbation-based evaluation, the degradation patterns we observe would be reduced to a single aggregate accuracy number. GUI-Perturbed reveals not just that training interventions degrade performance, but \emph{which} capability axes degrade and by how much. As the field moves toward more complex post-training recipes (multi-stage SFT, RL from grounding feedback, process rewards), evaluation tools that provide this level of diagnostic granularity will be essential for measuring whether new methods are making progress on the capabilities that matter.

\section{Limitations}
\label{sec:limitations}

\paragraph{Dataset scope.} The evaluation set contains 390 samples per variant, covers web-only scenarios, and is sourced from Mind2Web. This is a deliberate choice for controlled evaluation; cross-platform (desktop, mobile) and larger-scale extensions are future work.

\paragraph{Model coverage.} All open models share a single architecture lineage (Qwen2.5VL-7B). This design isolates the effect of post-training but limits generalization to other architectures and scales. We do not evaluate frontier commercial CUAs in this work.

\paragraph{Training recipe.} Training experiments use LoRA rank 8 only, a conservative configuration rather than the full space of post-training methods. We do not claim that training cannot address these failures; we claim that the default recipe does not, and that standard benchmarks alone cannot diagnose which capability axes are affected.

\paragraph{Perturbation realism.} Some perturbations produce visual outputs that no production website would generate. We prioritize diagnostic coverage over photo-realism: a perturbation that reveals a model's reliance on color as a grounding cue is informative regardless of whether the specific color combination is realistic.

\paragraph{Evaluation coverage.} Functional alignment is not tested in isolation. Instruction diversity covers a subset of natural referring expressions; colloquial and ambiguous references are left for future work.

\section{Future Work}
\label{sec:future}

\paragraph{Behavior-driven data curation.} Our results suggest that training data diversity along cognitive behavioral axes (spatial reasoning, instruction disambiguation, visual invariance) matters more than surface-level diversity (more platforms, more applications). Developing data curation pipelines organized around the capabilities they exercise, rather than the screenshots they contain, is a promising direction for improving CUA robustness.

\paragraph{Richer post-training recipes.} Rank-8 LoRA SFT with cross-entropy loss is insufficient for spatial grounding alignment. Multi-stage approaches that combine SFT with reinforcement learning, as explored in SpatialLadder~\citep{li2025spatialladder} and GuirlVG~\citep{kang2025guirlvg}, and process reward models that provide step-level supervision for grounding decisions rather than sequence-level loss, may offer a path forward.

\paragraph{Environment state representations.} Current GUI training operates on a static mapping from screenshot and instruction to action. Incorporating next-state feedback, i.e., the result of taking an action, would enable richer credit assignment and more efficient learning. Building environment representations that go beyond static screenshots is an active area of our research.

\paragraph{Extended coverage.} GUI-Perturbed currently covers web-only scenarios with a subset of natural referring expressions. Extending to desktop and mobile platforms, and expanding instruction diversity to include colloquial and ambiguous references, would broaden the diagnostic coverage of the benchmark.

\section{Conclusion}
\label{sec:conclusion}

We have presented GUI-Perturbed, a controlled perturbation framework that applies domain randomization to GUI grounding evaluation. By varying visual scenes and instructions along independent, controlled axes, GUI-Perturbed reveals three classes of brittleness in current models: spatial reasoning is systematically deficient (27--56~pp accuracy collapse on relational instructions), learned visual heuristics are static (a 70\% zoom degrades models scoring over 85\% on standard benchmarks), and chain-of-thought reasoning is miscalibrated across task types. Training experiments with rank-8 LoRA SFT show that naive data augmentation does not close these gaps, while our baseline comparison across three post-training stages suggests that RL with grounding-specific reward is more effective than SFT at improving spatial robustness. GUI-Perturbed provides diagnostic granularity into which specific failure axes are affected, complementing the aggregate signal that standard benchmarks provide.

The GUI-DR pipeline, GUI-Perturbed dataset, UI-TARS-1.5-7B-GUI-Perturbed model checkpoint, and result viewers are publicly available.

\begin{ack}
The authors thank Fig for its fiscal support of this work, the Mind2Web team for releasing the MHTML archives that form the foundation of GUI-Perturbed, and the Qwen, UI-TARS, and GTA1 teams for open-sourcing their models, which allowed controlled evaluation in this work. 
\end{ack}

\bibliographystyle{unsrtnat}
\bibliography{references}

\newpage
\appendix

\section{Failure Mode Taxonomy}
\label{sec:appendix:taxonomy}

We present qualitative examples for each failure mode identified in \cref{tab:failure-modes}. Each example includes the instruction, model output, and annotated screenshot.


\subsection*{Spatial Failures}

\begin{figure}[H]
\centering
\includegraphics[width=0.85\linewidth]{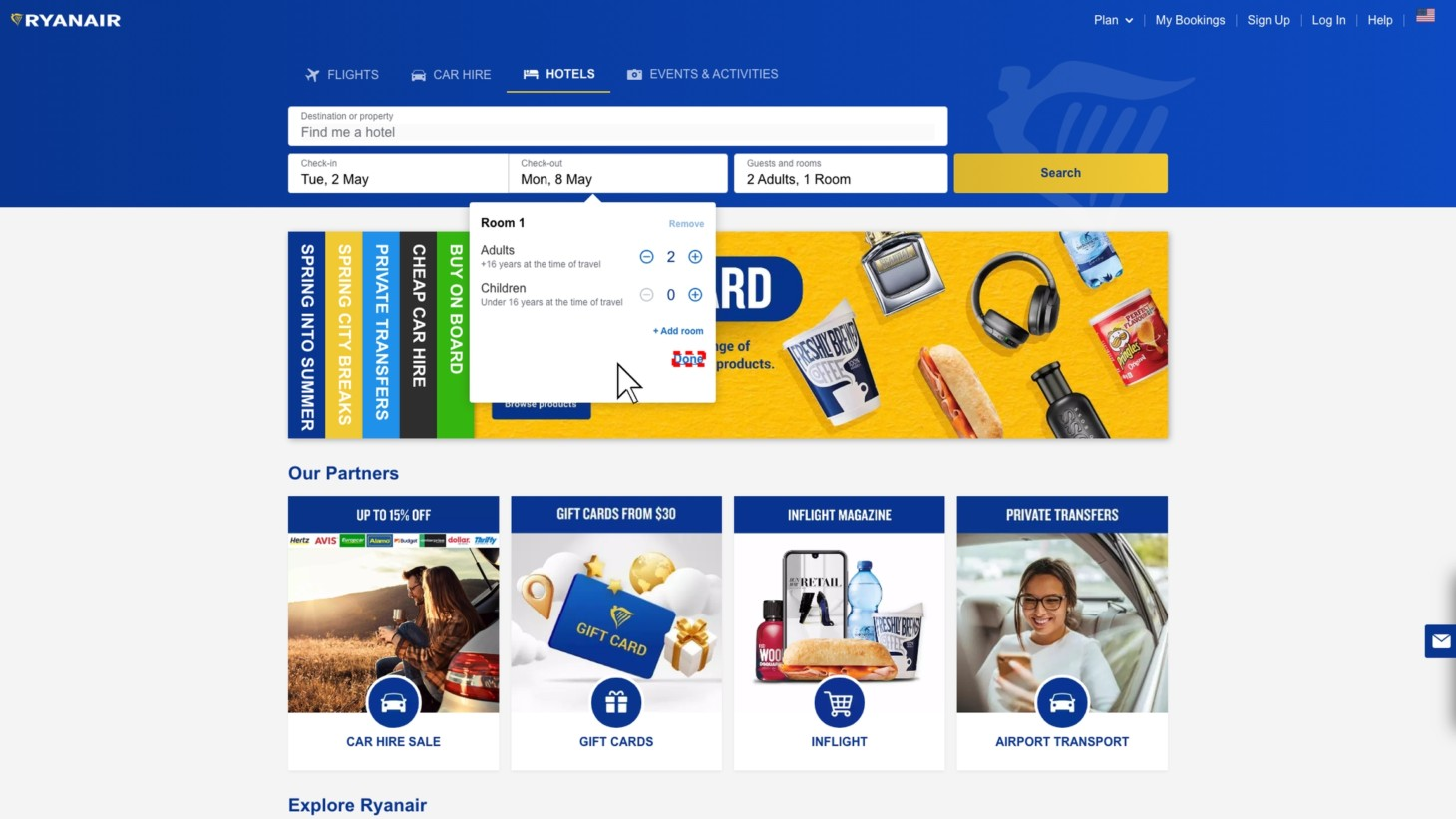}
\caption{\textbf{Click Region Error.} The model selects the correct UI element conceptually but clicks the wrong physical area of it. \emph{Instruction:} ``Click on `Done' button.'' \emph{Model output:} \texttt{click(start\_box='(639,438)')}.}
\label{fig:fm-click-region}
\end{figure}

\begin{figure}[H]
\centering
\includegraphics[width=0.85\linewidth]{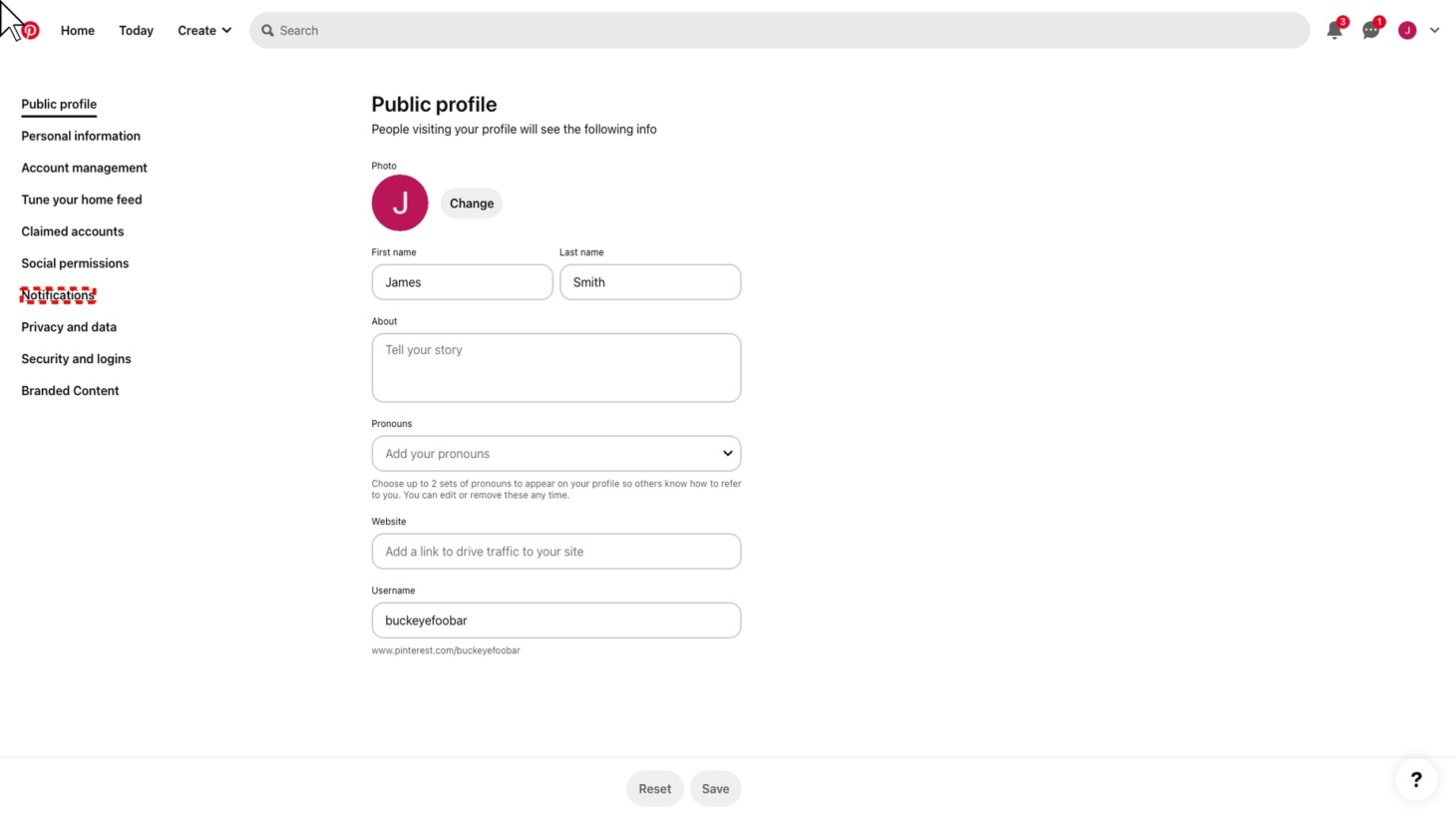}
\caption{\textbf{Location Hallucination.} The model correctly identifies what to click but fabricates or misplaces its on-screen coordinates. \emph{Instruction:} ``Click on `Notifications' div.'' \emph{Model output:} ``Thought: I noticed that there is a `Notifications' option in the left sidebar\ldots This option is located just below `Privacy and data' and above `Security and logins.'\,'' The model's reasoning is correct but the predicted coordinates do not correspond to the described element.}
\label{fig:fm-location-hallucination}
\end{figure}

\begin{figure}[H]
\centering
\includegraphics[width=0.85\linewidth]{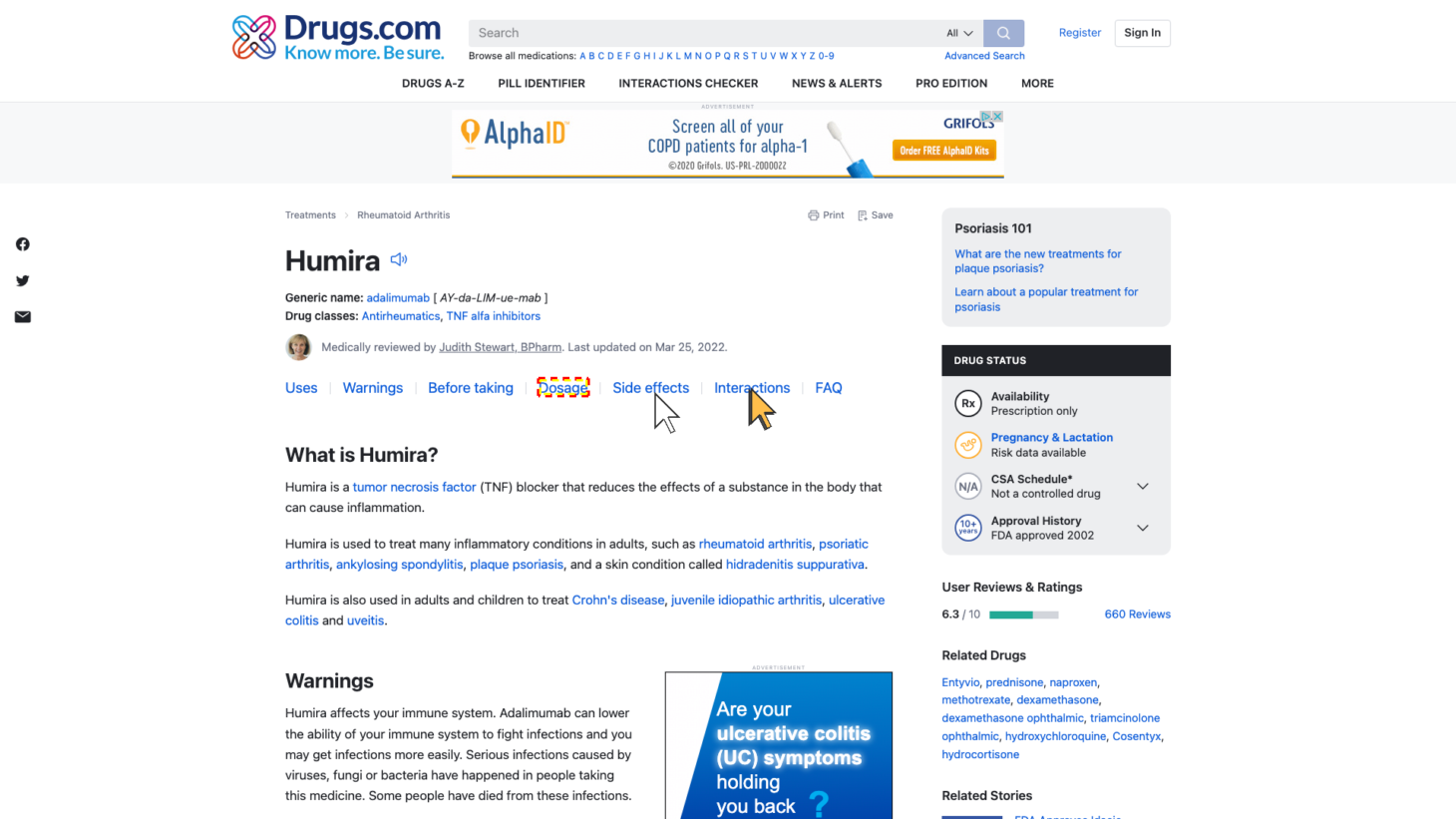}
\caption{\textbf{Spatial Reasoning Error.} The model incorrectly interprets relative spatial relationships. \emph{Instruction:} ``Click on the link to the left of `Side effects'.'' \emph{Model output:} ``Thought: I noticed that there is a link labeled `Interactions' located to the left of `Side effects'\ldots'' The model correctly names the target but clicks a link on the \emph{right} instead of the left.}
\label{fig:fm-spatial-reasoning}
\end{figure}


\subsection*{Semantic Failures}

\begin{figure}[H]
\centering
\includegraphics[width=0.85\linewidth]{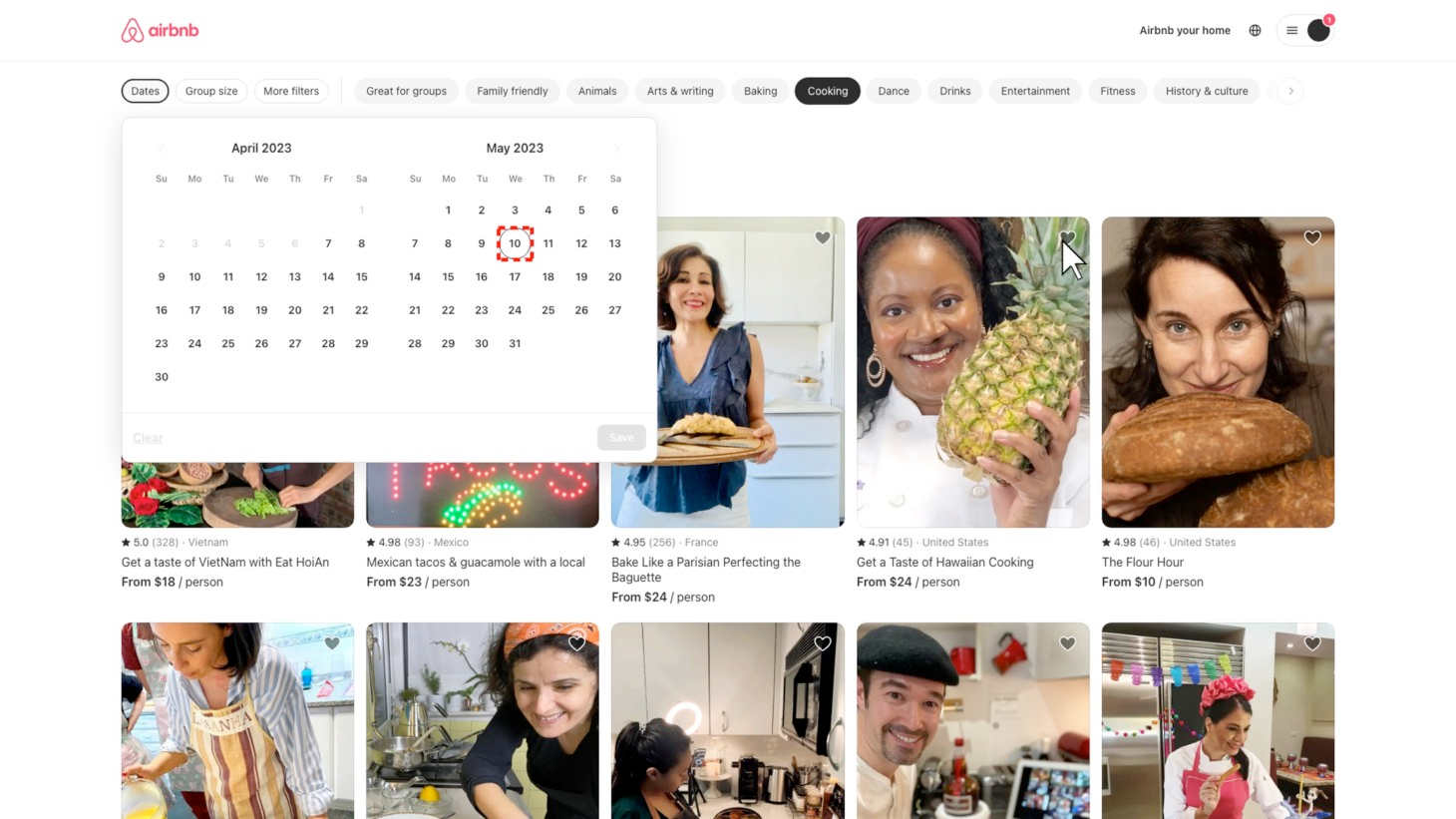}
\caption{\textbf{Goal Hallucination.} The model invents user intentions not specified in the instruction. \emph{Instruction:} ``Click on the button above `May 17th 2023'.'' \emph{Model output:} ``Thought: I noticed that there is a gray heart-shaped button\ldots clicking on it should allow me to save this cooking activity to my favorites.'' The ``save to favorites'' goal is absent from the instruction.}
\label{fig:fm-goal-hallucination}
\end{figure}

\begin{figure}[H]
\centering
\includegraphics[width=0.85\linewidth]{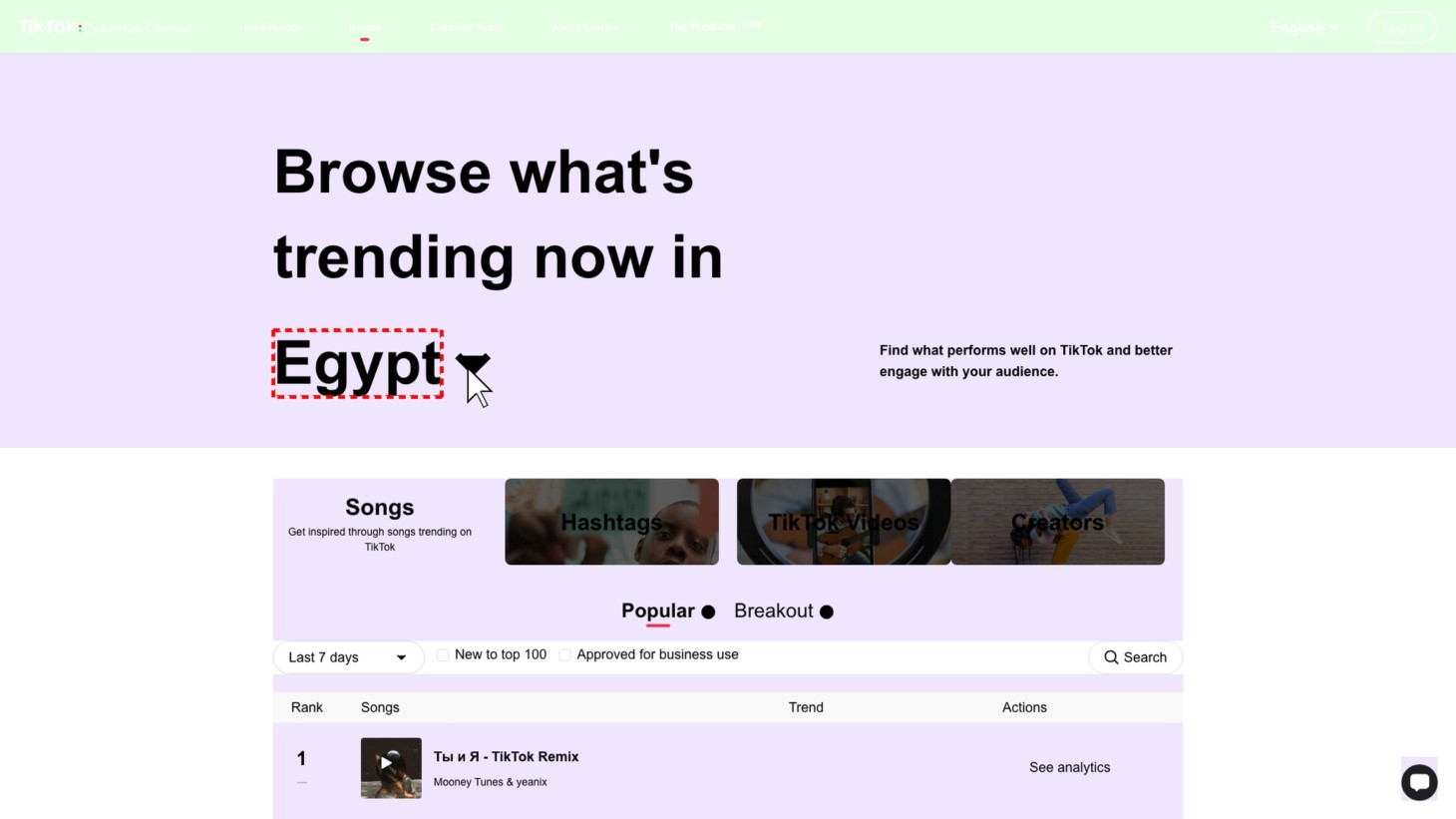}
\caption{\textbf{Instruction Misinterpretation.} The model misunderstands what the instruction refers to and selects a related but incorrect element. \emph{Instruction:} ``Click on `Egypt'.'' \emph{Model output:} ``Thought: \ldots you need to click on the dropdown button located to the right of the word `Egypt.'\,'' The model reinterprets ``click on Egypt'' as ``click the dropdown arrow beside Egypt.''}
\label{fig:fm-instruction-misinterpretation}
\end{figure}

\begin{figure}[H]
\centering
\includegraphics[width=0.85\linewidth]{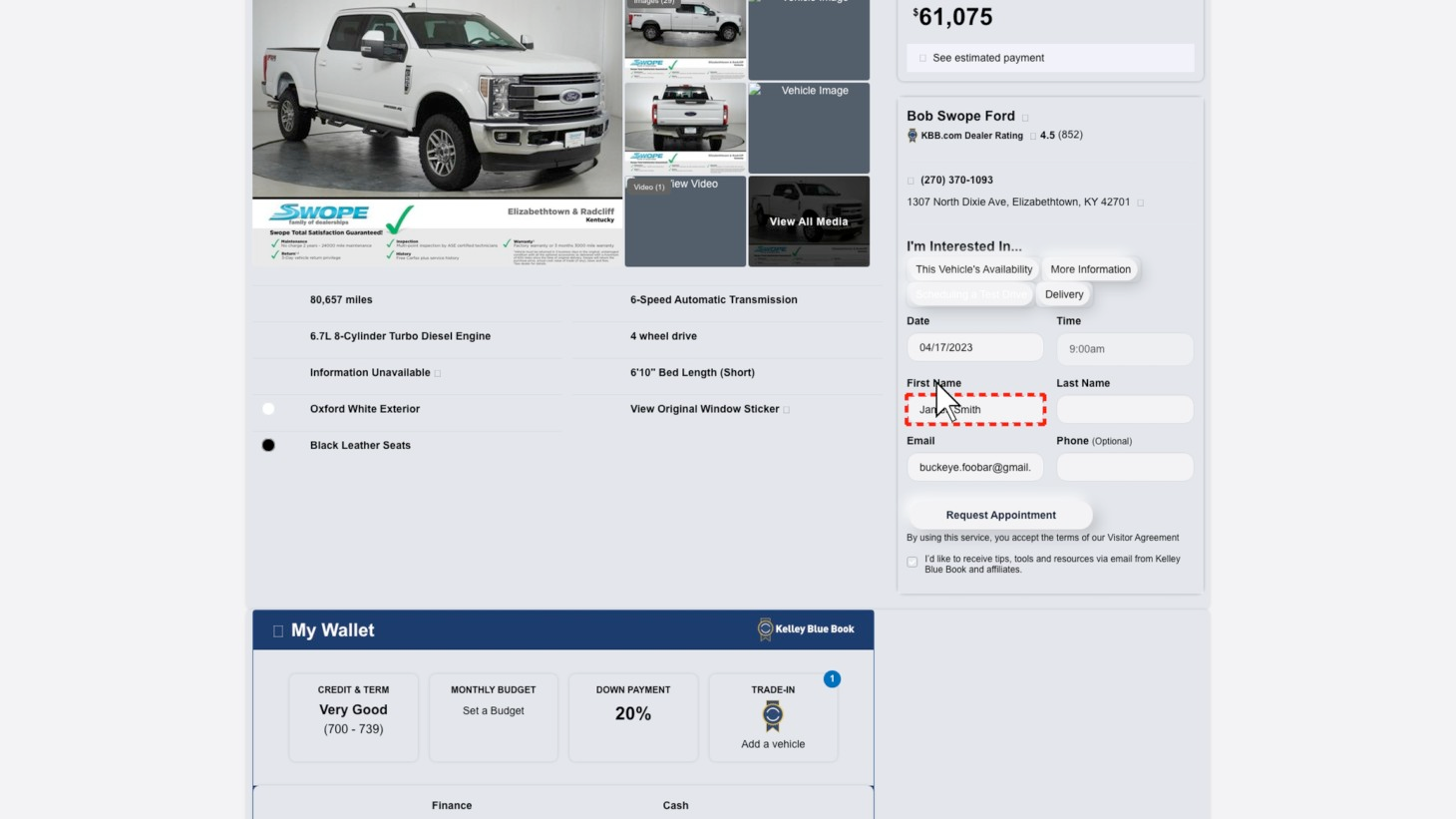}
\caption{\textbf{Text Matching Bias.} The model interacts with visible text matching the instruction without properly grounding it to the correct UI element. \emph{Instruction:} ``Click on `First Name' textbox.'' \emph{Model output:} \texttt{click(start\_box='(1242,509)')}. The model clicks the ``First Name'' label text rather than the input field beneath it.}
\label{fig:fm-text-matching}
\end{figure}


\subsection*{Visual Failure}

\begin{figure}[H]
\centering
\includegraphics[width=0.85\linewidth]{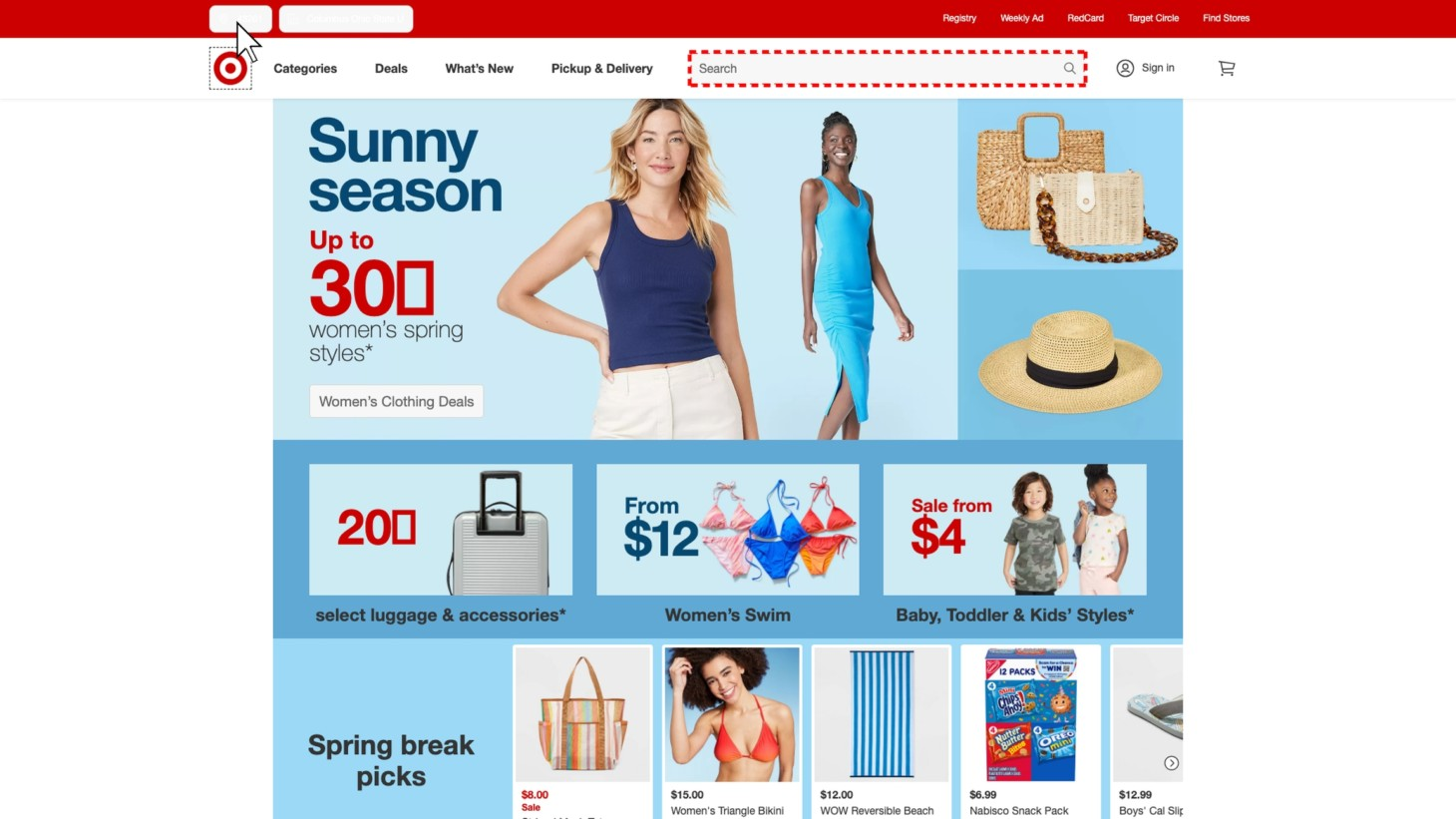}
\caption{\textbf{Visual Confusion.} The model relies on superficial visual cues (shape, color, position) and misidentifies the functional element. \emph{Instruction:} ``Type `bed sheets queen' in `Search: suggestions appear below' searchbox.'' \emph{Model output:} ``Thought: I noticed that there are two search boxes at the top of the page. The one on the left is the one I need.'' The model mistakes a light-colored button with faint text for the target search box.}
\label{fig:fm-visual-confusion}
\end{figure}


\subsection*{Reasoning Failure}

\begin{figure}[H]
\centering
\includegraphics[width=0.85\linewidth]{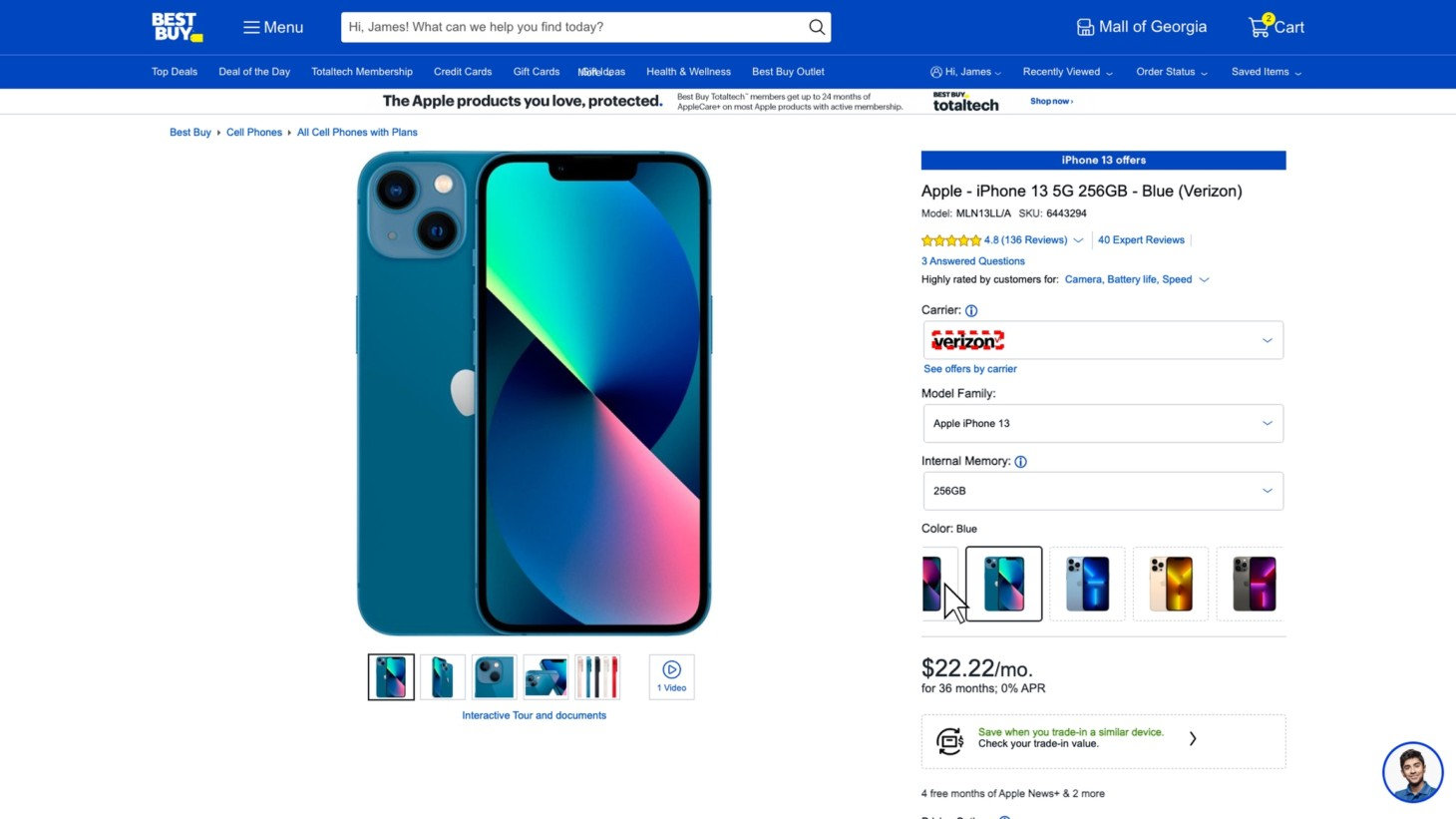}
\caption{\textbf{Reasoning Drift.} Adding explicit reasoning causes the model's grounded action prediction to deteriorate. \emph{Instruction:} ``Click on `Verizon' img.'' \emph{Model output:} ``Thought: The task is to click on the `Verizon' image\ldots The image is located at the bottom of the product details section.'' The phrase ``located at the bottom'' in the model's own reasoning misleads it to click an unrelated image at the bottom of the screen.}
\label{fig:fm-reasoning-drift}
\end{figure}

\section{Additional Metrics}
\label{sec:appendix:metrics}

We additionally compute bounding box center MSE ($\text{MSE} = \frac{1}{N}\sum_{i=1}^{N} \|\hat{p}_i - p_i\|_2^2$), normalized MSE ($\text{NMSE} = \frac{1}{N}\sum_{i=1}^{N} \frac{\|\hat{p}_i - p_i\|_2^2}{w_i \cdot h_i}$), and normalized distance ($D_{\text{norm}} = \frac{1}{N}\sum_{i=1}^{N} \frac{\|\hat{p}_i - p_i\|_2}{\sqrt{w_i^2 + h_i^2}}$). These metrics capture error magnitude but did not reveal trends beyond what hit rate and flip rate already capture in our experiments.

\section{Finetuned Model Robustness}
\label{sec:appendix:finetuned}

\begin{table}[t]
\centering
\caption{Perturbation robustness of finetuned models ($n = 390$ matched sample pairs per test). Same metrics and notation as \cref{tab:robustness-baseline}.}
\label{tab:robustness-finetuned}
\scriptsize
\begin{tabular}{llc cc cc cc}
\toprule
 & & & \multicolumn{2}{c}{\textbf{Flip Rate}} & \multicolumn{2}{c}{\textbf{Net $\Delta$ (\%)}} & & \\
\cmidrule(lr){4-5} \cmidrule(lr){6-7}
\textbf{Model} & \textbf{Pert.} & \textbf{Base Acc.} & \textbf{Dir.} & \textbf{Rel.} & \textbf{Dir.} & \textbf{Rel.} & $\boldsymbol{b}$\,/\,$\boldsymbol{c}$ & \textbf{Sig.} \\
\midrule
UI-TARS-1.5 (base) & Precision & 63.3 & 11.8\% & 18.7\% & +5.6*** & +5.4** & 162/76 & 4/4 \\
 & Style &  & 11.2\% & 19.1\% & +2.7 & +1.2 & 133/103 & 0/4 \\
 & Text Shrink &  & 6.5\% & 14.2\% & -0.9 & -0.1 & 77/85 & 0/4 \\
\midrule
FT-All (6.5k) & Precision & 63.1 & 12.7\% & 19.7\% & +5.5** & +5.6** & 170/83 & 4/4 \\
 & Style &  & 14.1\% & 18.7\% & +2.8 & +1.3 & 144/112 & 0/4 \\
 & Text Shrink &  & 6.9\% & 13.5\% & +1.0 & +2.4 & 93/66 & 0/4 \\
\midrule
FT-Style (6.5k) & Precision & 63.1 & 12.9\% & 19.0\% & +5.8** & +5.6** & 169/80 & 4/4 \\
 & Style &  & 14.0\% & 18.1\% & +2.9 & +1.4 & 142/108 & 0/4 \\
 & Text Shrink &  & 6.9\% & 12.9\% & +1.0 & +2.2 & 90/65 & 0/4 \\
\midrule
FT-TextShrink (6.5k) & Precision & 63.1 & 12.9\% & 19.6\% & +5.8** & +6.0** & 173/81 & 4/4 \\
 & Style &  & 13.8\% & 18.8\% & +2.8 & +1.7 & 145/110 & 0/4 \\
 & Text Shrink &  & 6.8\% & 12.9\% & +0.9 & +2.2 & 89/65 & 0/4 \\
\midrule
FT-All (25k, 3ep) & Precision & 61.4 & 13.1\% & 21.7\% & +5.9*** & +5.8** & 181/90 & 3/4 \\
 & Style &  & 15.6\% & 21.4\% & +4.4** & +0.6 & 164/125 & 1/4 \\
 & Text Shrink &  & 8.7\% & 14.9\% & +1.5 & +1.3 & 103/81 & 0/4 \\
\midrule
FT-Salesforce (25k) & Precision & 62.9 & 12.7\% & 19.5\% & +5.8** & +5.4* & 169/82 & 4/4 \\
 & Style &  & 13.8\% & 18.8\% & +2.6 & +1.2 & 142/113 & 0/4 \\
 & Text Shrink &  & 6.8\% & 12.3\% & +0.6 & +2.3 & 86/63 & 0/4 \\
\midrule
FT-Perturbed (25k) & Precision & 62.9 & 13.2\% & 19.5\% & +5.5** & +5.4* & 170/85 & 4/4 \\
 & Style &  & 13.8\% & 18.6\% & +2.8 & +1.2 & 142/111 & 0/4 \\
 & Text Shrink &  & 6.7\% & 12.9\% & +0.8 & +1.9 & 87/66 & 0/4 \\
\bottomrule
\end{tabular}
\end{table}

After finetuning, precision perturbation remained the primary source of significant degradation (27/28 tests significant), while style (1/28) and text-shrink (0/28) perturbations remained non-significant. None of the finetuning strategies substantially reduced the precision vulnerability. The $b$/$c$ ratios for precision remained close to 2:1 across all finetuned variants, indicating that the directional nature of the degradation persisted. Flip rates for style perturbation were comparable to or slightly higher than the baseline, suggesting finetuning did not improve prediction consistency.

\section{Evaluation Prompt Templates}
\label{sec:appendix:prompts}

We evaluate each model using its native prompt format in both reasoning (with chain-of-thought) and no-reasoning (direct action) configurations. All models receive a single screenshot resized via the Qwen2.5-VL smart resize algorithm (factor=28, min 100$\times$28$^2$ pixels, max 16384$\times$28$^2$ pixels). Below we summarize the prompt structure for each model.

\subsection*{UI-TARS-1.5-7B}

UI-TARS-1.5 uses a structured action space with bounding box coordinates. The system prompt is ``You are a helpful assistant.'' The user message contains the task instruction and screenshot.

\paragraph{With reasoning.} The model is instructed to output a \texttt{Thought} followed by an \texttt{Action}:

\begin{lstlisting}
## Output Format
Thought: ...
Action: ...

## Action Space
click(start_box='<|box_start|>(x1,y1)<|box_end|>')
left_double(start_box='<|box_start|>(x1,y1)<|box_end|>')
right_single(start_box='<|box_start|>(x1,y1)<|box_end|>')
drag(start_box='...', end_box='...')
hotkey(key='')
type(content='')
scroll(start_box='...', direction='down|up|right|left')
wait()
finished()
call_user()

## Note
- Use English in Thought part.
- Write a small plan and summarize your next action
  in one sentence in Thought part.
\end{lstlisting}

\paragraph{Without reasoning.} The same action space but the output format omits the \texttt{Thought} field, requesting only \texttt{Action: ...} directly.

\subsection*{GTA1-7B}

GTA1 uses a coordinate-only output format. The system prompt specifies the image resolution and requests a single $(x,y)$ point prediction.

\paragraph{With reasoning.}
\begin{lstlisting}
You are an expert UI element locator. Given a GUI image
and a user's element description, provide the coordinates
of the specified element as a single (x,y) point.
The image resolution is height {h} and width {w}.
For elements with area, return the center point.

## Output Format
Thought: ...
Action: (x,y)
\end{lstlisting}

\paragraph{Without reasoning.} The same system prompt but the output format requests only \texttt{(x,y)} without a \texttt{Thought} field.

\subsection*{Qwen2.5-VL-7B}

Qwen2.5-VL uses a tool-calling format with a \texttt{computer\_use} function. The system prompt defines the full action space as a JSON function signature, including actions: \texttt{key}, \texttt{type}, \texttt{mouse\_move}, \texttt{left\_click}, \texttt{left\_click\_drag}, \texttt{right\_click}, \texttt{middle\_click}, \texttt{double\_click}, \texttt{scroll}, \texttt{wait}, and \texttt{terminate}. The screen resolution is injected dynamically based on the resized image dimensions.

\paragraph{With reasoning.} The system prompt prepends an output format section:
\begin{lstlisting}
# Output Format
Before making a tool call, you should think through
your approach. Use the following format:

Thought: [Write a small plan analyzing the current
screenshot, identifying the target element(s), and
summarizing your next action in one sentence.]

Then make your tool call.
\end{lstlisting}

\paragraph{Without reasoning.} The tool-calling format is used directly without the \texttt{Thought} prefix.

\subsection*{Image Preprocessing}

All screenshots are preprocessed using the Qwen2.5-VL smart resize algorithm before being passed to each model. The algorithm enforces divisibility by a factor of 28 while respecting minimum and maximum pixel budgets, and caps the aspect ratio at 200:1. Images are encoded as base64 PNG and passed via the \texttt{image\_url} field in the OpenAI-compatible message format.

\end{document}